\documentclass[conference,letterpaper]{IEEEtran}

\usepackage[utf8]{inputenc}
\usepackage[T1]{fontenc}
\usepackage{graphicx}
\usepackage{amsmath,amssymb,amsfonts}
\usepackage{amsthm}
\newtheorem{proposition}{Proposition}
\newtheorem*{proposition*}{Proposition}
\usepackage{booktabs,multirow}
\usepackage{algorithm}
\usepackage{algorithmic}
\usepackage[numbers,sort&compress]{natbib}
\usepackage{nicefrac}
\usepackage{microtype}
\usepackage{xcolor}
\newif\ifshowrevisions
\showrevisionsfalse
\newcommand{\revisioncolor}{\ifshowrevisions\color{blue}\fi}
\newcommand{\paperrev}[1]{{\revisioncolor#1}}
\let\theoryrev\paperrev
\newenvironment{theoryrevision}{\par\begingroup\revisioncolor}{\par\endgroup}
\usepackage{needspace}
\usepackage{url}
\usepackage[hidelinks]{hyperref}
\hypersetup{
  pdftitle={Flow-CAD: Flow Matching for Contextual Anomaly Detection and Explanation in Tabular Data},
  pdfauthor={Zhong Li, Qi Huang, Yuhang Wang, Yang Cao, Thomas Bäck, Niki van Stein, Matthijs van Leeuwen, Lincen Yang},
  pdfsubject={Contextual anomaly detection in tabular data}
}

\title{Flow-CAD: Flow Matching for Contextual Anomaly Detection and Explanation in Tabular Data}

\renewcommand{\IEEEauthorrefmark}[1]{\raisebox{0pt}[0pt][0pt]{\textsuperscript{\footnotesize #1}}}

\author{
\IEEEauthorblockN{
Zhong Li\paperrev{\IEEEauthorrefmark{1,2}},
Qi Huang\paperrev{\IEEEauthorrefmark{2}},
Yuhang Wang\paperrev{\IEEEauthorrefmark{3}},
\paperrev{Yang Cao\IEEEauthorrefmark{1}},\\
Thomas B\"ack\paperrev{\IEEEauthorrefmark{2}},
Niki van Stein\paperrev{\IEEEauthorrefmark{2}},
\paperrev{Matthijs van Leeuwen\IEEEauthorrefmark{2}},
Lincen Yang\paperrev{\IEEEauthorrefmark{2,*}}}
\IEEEauthorblockA{\paperrev{\IEEEauthorrefmark{1}}Great Bay University, Dongguan, China}
\IEEEauthorblockA{\paperrev{\IEEEauthorrefmark{2}Leiden University, Leiden, The Netherlands}}
\IEEEauthorblockA{\paperrev{\IEEEauthorrefmark{3}KU Leuven, Leuven, Belgium}}
\IEEEauthorblockA{\paperrev{\IEEEauthorrefmark{*}Corresponding author: Lincen Yang (\texttt{l.yang@liacs.leidenuniv.nl})}}
}

\begin{document}

\maketitle
\thispagestyle{plain}
\pagestyle{plain}

\begin{abstract}
Contextual anomaly detection in tabular data aims to identify objects whose behavior is unexpected under their contextual conditions. \paperrev{In data-engineering pipelines, this task screens table rows whose measurements violate context-dependent integrity relationships.} Most existing approaches rely on either shallow conditional models with limited distributional expressiveness or localized estimators that require per-context modeling, making it difficult to capture complex conditional relationships and scale to high-dimensional settings. We propose Flow-CAD, a new framework for contextual anomaly detection based on Flow Matching. Flow-CAD learns a single global and amortized conditional generative model that flexibly represents behavioral distributions conditioned on context, without assuming simple parametric forms or retraining local models. To enable robust and interpretable anomaly scoring, we introduce a scale-normalized and truncated residual energy derived from the learned conditional flow, which stabilizes anomaly scores, mitigates domination by extreme deviations, and supports feature-wise attribution. We further provide theoretical insights showing that the proposed residual-based anomaly score \theoryrev{is stable under velocity estimation error}, and that its Monte Carlo estimator \theoryrev{satisfies concentration bounds}. Experiments on 20 benchmark datasets with 13 competitors demonstrate that Flow-CAD \paperrev{has the best average rank for PR AUC, ROC AUC, and P@$n$}, highlighting its effectiveness and practicality for contextual anomaly detection in tabular data. We provide the source~code~at:~\paperrev{\url{https://github.com/ZhongLIFR/FlowCAD}}.
\end{abstract}

\section{Introduction}

Contextual anomaly detection aims to identify data instances whose behavior is
abnormal when interpreted relative to their contextual conditions \citep{song2007conditional,li2023explainable}. Under this setting, features are typically partitioned into contextual variables (e.g., operating conditions, demographic attributes, or environmental factors) and behavioral variables (e.g., measurements, responses, or outcomes). An instance is considered anomalous not because it is globally rare, but because its behavior is unlikely given its context. This setting arises naturally in a wide range of applications, including industrial monitoring \citep{kayan2024casper}, environmental analysis \citep{hayes2015contextual}, and security in smart grids \citep{kosek2016contextual}, where identical behaviors may be normal in one context but anomalous in another. \paperrev{For example, a plant may record its operating mode alongside pressure and temperature readings in each row. Each reading can be typical for that mode on its own, while the combination is unusual. In database-backed analytics, screening for such rows can precede aggregation, monitoring, and model~training~\citep{ding2025cbaclean,siddiqi2023saga}. For a flagged row, Flow-CAD reports how much each behavioral column contributes to the anomaly score, helping analysts decide which~measurements~to~inspect~\citep{schelter2018quality}.}

\begin{figure}[!ht]
    \centering
    \includegraphics[width=\columnwidth]{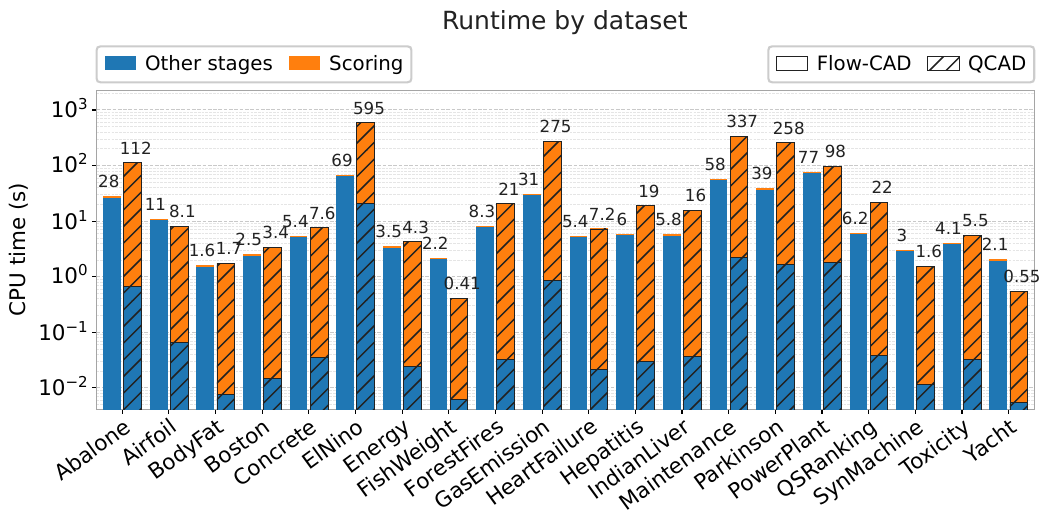}
    \caption{Runtime comparison between \textbf{Flow-CAD} and \textbf{QCAD} \paperrev{on all $20$ benchmark datasets (data seed $42$, one Apple M4 CPU thread per run). Orange shows the time to compute row scores and feature contributions. Blue includes data loading and metric computation for both methods, plus the $1{,}000$-epoch model fit for Flow-CAD and context encoding and Gower-neighborhood construction for QCAD.} Flow-CAD is shown with solid bars and QCAD with striped bars. \paperrev{The vertical axis is logarithmic. Bar labels show total CPU time.}}
    \label{fig:runtime_comparison}
\end{figure}

Despite steady progress, existing contextual anomaly detection methods continue
to face several fundamental limitations when applied to complex tabular data.
Early generic frameworks such as CAD~\citep{song2007conditional} rely on parametric assumptions and hand-crafted mapping functions to capture dependencies between contextual and behavioral variables, which restrict flexibility, scalability, and support for mixed feature types. A subsequent approach ROCOD~\citep{liang2016robust} \paperrev{combines} global
regression models with local neighborhood statistics, but primarily \paperrev{characterizes} conditional behavior through expectations, which can be inadequate when the conditional distribution is skewed, multimodal, or context-dependent. A recent method QCAD~\citep{li2023explainable} improves robustness and interpretability by estimating conditional quantiles, yet remains fundamentally localized, relying on per-query neighborhood construction or local conditional estimation. \paperrev{This repeated local estimation adds scoring overhead (see Figure~\ref{fig:runtime_comparison}).} Together, these limitations motivate a global and amortized approach that avoids repeated local modeling, captures complex context-dependent behavior, and supports efficient inference with interpretable anomaly scores.

Recent advances in deep generative modeling \paperrev{\citep{wang2024challenges,li2026diffusion}} provide an opportunity to revisit
contextual anomaly detection from a global and amortized perspective.
In principle, conditional generative models are well suited for representing
complex conditional distributions without relying on restrictive parametric
assumptions. \paperrev{For continuous flows, likelihood evaluation typically requires numerical integration of an ordinary differential equation (ODE) and estimation of the velocity field's divergence. Likelihood values can also be unreliable as a direct anomaly signal in complex distributions \citep{kamkari2024geometric,nalisnick2019deep}.} As a result, existing generative approaches do not naturally satisfy the combined requirements of efficient inference, robust anomaly scoring, and interpretability in contextual anomaly detection. Flow Matching~\citep{lipman2023flow,albergo2023building} offers an
alternative training paradigm for continuous generative models that avoids
explicit likelihood evaluation and ODE-based density computation \paperrev{during training}.
By learning a conditional velocity field that transports samples from a simple
base distribution to the data, Flow Matching retains the expressive power of
conditional generative models while enabling path-wise, residual-based
assessments of how well observed behaviors are explained under a given context.
Motivated by these properties, we propose Flow-CAD, a new framework for contextual anomaly detection based on Flow Matching.

Specifically, Flow-CAD learns a single global and amortized conditional generative model that captures the \paperrev{joint} distribution of behavioral variables conditioned on context. Instead of relying on explicit likelihood evaluation, Flow-CAD derives anomaly scores from residuals of the learned conditional flow, measuring how well the model explains an observed behavior given its context. To obtain robust and interpretable anomaly scores, we further introduce a scale-normalized and truncated residual energy that aggregates deviations across behavioral dimensions while \paperrev{limiting the influence of extreme values}. This construction yields native, decomposable feature-level attributions, which \paperrev{identify the behavioral columns contributing to a flagged row's score}. \theoryrev{In addition, we relate the ideal-field residual to conditional denoising, establish the stability of the normalized and truncated score under velocity estimation error, and analyze the variance and concentration of its nested~Monte~Carlo~estimator.}

\textbf{Contributions.}
The main contributions of this work are as follows.
(1)~We introduce Flow-CAD, \paperrev{a} contextual anomaly detection framework based on Flow Matching \paperrev{for screening database tables. A single model learns how the behavioral columns vary together given the context columns.}
(2)~We propose a residual-based anomaly energy for conditional flows \paperrev{that scores rows in batches without computing likelihoods or refitting for each query. A data engineering pipeline can use these scores to flag records before aggregation or model~training.}
(3)~We develop a normalized and truncated aggregation scheme that yields robust scores and \paperrev{additive column contributions. For a flagged row, these contributions show which columns account for most of its score, helping analysts decide which values to inspect~first.}
(4)~We provide theoretical analyses \paperrev{and compare Flow-CAD with $13$ baselines on $20$ benchmark tables, using a fixed training budget for our model. In a separate test on held-out synthetic rows, we ask whether the model can detect a context mismatch that only becomes visible when three behavioral columns are considered together. We also compare runtime with QCAD on full tables and report batch scoring time and peak memory on synthetic inputs of up to one~million~rows.}

\section{Related Work}

\textbf{Contextual Anomaly Detection.}
Anomaly detection in tabular data has been extensively studied, with prior work commonly distinguishing between point, contextual, and collective anomalies~\paperrev{\citep{chandola2009anomaly,han2022adbench}}. This work focuses on \emph{contextual anomaly detection}, where an instance is considered anomalous if its behavior is unlikely given its contextual attributes. Early methods \paperrev{such as CAD~\citep{song2007conditional} use probabilistic mixtures}, while subsequent approaches explore probabilistic and correlation-based formulations such as ROCOD~\citep{liang2016robust} and LoPAD~\citep{lu2020lopad}.
More recent methods emphasize interpretability, with QCAD~\citep{li2023explainable} modeling conditional distributions via quantile regression forests and deriving feature-level explanations from conditional deviations. \paperrev{RFOD~\citep{ang2026rfod} predicts each feature from the remaining features, and Normalcy Score~\citep{bindini2026dealing} quantifies uncertainty for a scalar behavioral target. Local models and per-context estimation can incur repeated computation at scoring time.}

\textbf{Generative Models for Anomaly Detection.}
\paperrev{Generative approaches to anomaly detection often use low likelihood as a signal of abnormality.} Classical density models and, more recently, deep generative approaches have been explored for tabular anomaly detection~\paperrev{\citep{pang2021deep,wang2024challenges,li2026diffusion}}. Normalizing flows further enable exact likelihood evaluation through invertible transformations~\citep{rezende2015variational}, but their practical application is often constrained by architectural design and the cost of likelihood computation in high-dimensional settings. Flow Matching~\paperrev{\citep{lipman2023flow,albergo2023building,liu2023flow,tong2024minibatch}} has recently emerged as a scalable alternative for training continuous normalizing flows without explicit likelihood or ODE-based density evaluation. Building on this paradigm, \paperrev{TCCM~\citep{li2025scalable} uses a single contraction residual toward the origin for \emph{point} anomaly detection. Flow Mismatching~\citep{chen2026flowmismatching} aggregates residuals over Gaussian-to-image paths for visual defects, fitting on normal images. On contaminated tables, Konz et al.~\citep{konz2026unsupervised} compare TCCM and Forest-Flow, whose Deviation score also pools velocity residuals over time and noise. Flow-CAD conditions the joint behavioral field on separate context columns and uses an additive score with median absolute deviation (MAD) scaling and clipping.}

\textbf{Explainability.}
Explainability is increasingly recognized as a critical requirement in anomaly detection, particularly for tabular data where decisions often have high practical impact.
Recent surveys~\citep{yepmo2022anomaly,li2023survey} distinguish between post-hoc and intrinsic explanation methods, and highlight the importance of feature-level attribution in practical deployments. Flow-CAD contributes to this line of work by providing \emph{intrinsic} and \emph{decomposable} anomaly explanations derived directly from the \paperrev{coordinate terms of its residual score}, without relying on post-hoc explanation models.

\section{Flow-CAD: Conditional Flow-Based Contextual Anomaly Detection}
\label{sec:flowcad}

\begin{figure*}[tp]
    \centering
    \includegraphics[width=0.99\linewidth]{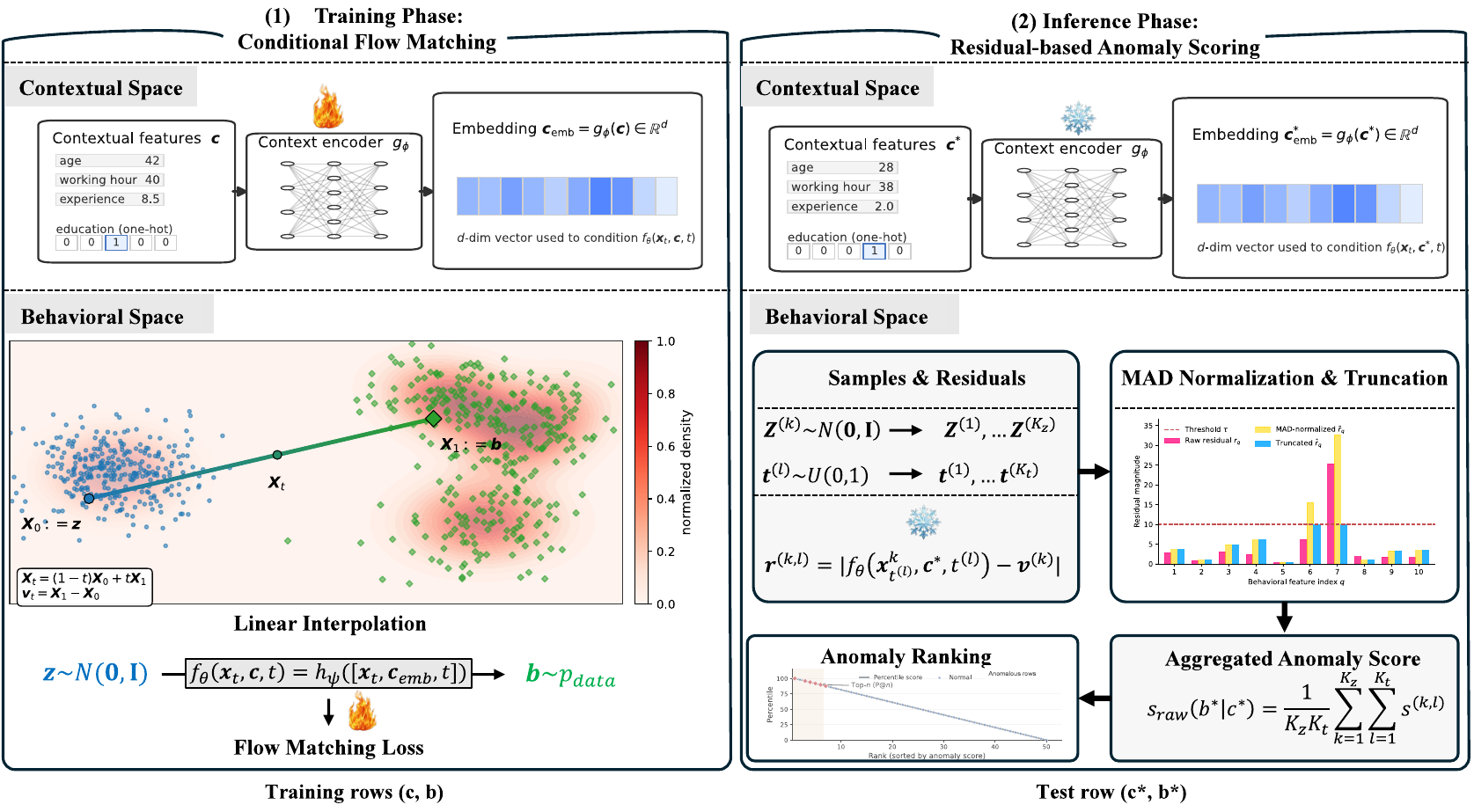}
    \caption{Illustration of the training (left) and inference (right) phases of Flow-CAD. \paperrev{The inference panel writes $t^{(\ell)}$ for $t^{(k,\ell)}$ within~Gaussian~block~$k$.}}
    \label{fig:workflow}
\end{figure*}

\subsection{Problem Setup and Method Overview}
\label{sec:method-overview}

We study \emph{contextual anomaly detection} in tabular data. Each instance is represented by a pair $(\boldsymbol{c},\boldsymbol{b})$, where $\boldsymbol{c}\in\mathbb{R}^{P}$ denotes the \paperrev{encoded context vector} and
$\boldsymbol{b}\in\mathbb{R}^{Q}$ denotes behavioral features. \paperrev{Given a table $\{(\boldsymbol{c}_i,\boldsymbol{b}_i)\}_{i=1}^{N}$, our goal is to prioritize rows whose behavior conflicts with patterns expected under their context. We denote the corresponding conditional behavior distribution by $p(\boldsymbol{b}\mid\boldsymbol{c})$.}

\textbf{Flow Matching as a modeling primitive.}
\theoryrev{Flow Matching trains a continuous-time velocity field by regressing pathwise transport targets along prescribed interpolation paths~\citep{lipman2023flow,tong2024minibatch}. Its population optimum is the conditional mean target given the intermediate state, context, and time. The regression objective requires neither density evaluation nor ODE integration. Flow-CAD learns one \emph{conditional} velocity field $f_\theta(\boldsymbol{x}_t,\boldsymbol{c},t)$ for the behavior distribution $p(\boldsymbol{b}\mid\boldsymbol{c})$ and reuses \mbox{the field across contexts}.}

\textbf{Method Overview.}
Figure~\ref{fig:workflow} illustrates the overall pipeline of Flow-CAD, which consists of two phases: \textit{(1) Training Phase: Conditional Flow Matching} (Section~\ref{sec:fm}), in which
we train a context-conditioned neural velocity field via Flow Matching along a linear interpolation path between a Gaussian base sample and the observed behavioral vector. This yields a single global model that captures conditional behavior across all contexts; \textit{(2) Inference Phase: Residual-based contextual anomaly scoring} (Section~\ref{sec:scoring}), in which we quantify contextual abnormality by the discrepancy between the model-predicted velocity and the true velocity induced by an observed instance, aggregated over multiple noise and time samples. To ensure robustness and interpretability, residuals are normalized using the median absolute deviation (MAD) and truncated to mitigate the dictator effect where a single extreme deviation dominates the score. \paperrev{It also yields additive output-coordinate terms for inspecting flagged rows.}

\subsection{Phase 1: Training Detection Model using Flow Matching with Linear Interpolation}
\label{sec:fm}

Let $\boldsymbol{b} \in \mathbb{R}^Q$ denote behavioral features and $\boldsymbol{c} \in \mathbb{R}^P$ denote \paperrev{encoded} contextual features. We introduce a conditional velocity field
$
f_\theta : \mathbb{R}^Q \times \mathbb{R}^P \times [0,1] \to \mathbb{R}^Q,
\quad
(\boldsymbol{x}_t, \boldsymbol{c}, t) \mapsto f_\theta(\boldsymbol{x}_t,\boldsymbol{c},t),
$
parameterized by a neural network with learnable parameters $\theta$. This velocity field is trained via Flow Matching on a simple linear path \citep{liu2023flow} between a base distribution and the data in the behavioral space.

\textbf{Linear path.}
\paperrev{Let $p_0$ be the density of $\mathcal{N}(\boldsymbol{0}, I_Q)$, where $I_Q$ is the $Q$-dimensional identity. For $\boldsymbol{z}\sim p_0$ and $t\in[0,1]$, the path to $\boldsymbol{b}$ is $\boldsymbol{x}_t=(1-t)\boldsymbol{z}+t\boldsymbol{b}$.} \theoryrev{For each sampled endpoint--noise pair, the pathwise target velocity is}
$
  \boldsymbol{v}_t 
  = \frac{d \boldsymbol{x}_t}{dt}
  = \boldsymbol{b} - \boldsymbol{z},
$
which is independent of $t$.

\textbf{Flow Matching objective.}
We train $f_\theta$ so that it approximates this velocity along the path. \theoryrev{We write the Flow Matching objective in squared-norm form as}
\begin{equation}
  \mathcal{L}(\theta)
  =
  \mathbb{E}_{(\boldsymbol{b},\boldsymbol{c}),\,\boldsymbol{z},\,t}
  \left[
    \left\|
      f_\theta(\boldsymbol{x}_t,\boldsymbol{c},t)
      - (\boldsymbol{b} - \boldsymbol{z})
    \right\|_2^2
  \right],
  \label{eq:fm-loss}
\end{equation}
\paperrev{We estimate this expectation over minibatches, drawing $\boldsymbol z\sim p_0$ and $t\sim\mathrm{Unif}(0,1)$ independently for each row.} \theoryrev{The implementation averages over behavioral coordinates as well, multiplying this objective by $1/Q$ without changing~its~\mbox{minimizer}.}

\textbf{Context encoder and architecture.}
In practice, $f_\theta$ is implemented as a context-conditional multi-layer perceptron (MLP):
$
  \boldsymbol{c}_{\mathrm{emb}}
  = g_\phi(\boldsymbol{c}), $ and $
  f_\theta(\boldsymbol{x}_t,\boldsymbol{c},t)
  = h_\psi\bigl([\boldsymbol{x}_t,\ \boldsymbol{c}_{\mathrm{emb}},\ t]\bigr),
$
where $g_\phi$ and $h_\psi$ are MLPs, and $[\cdot,\cdot,\cdot]$ denotes concatenation. \paperrev{The embedding $\boldsymbol c_{\mathrm{emb}}\in\mathbb R^d$ has width $d$.}
Categorical contextual features are one-hot encoded and then fed into $g_\phi$ together with numerical contextual features, so that the model can handle mixed-type $\boldsymbol{c}$ in a unified way.
The parameters \paperrev{$\theta=(\phi,\psi)$} are optimized jointly via Eq. \eqref{eq:fm-loss}~using~Adam~\citep{kingma2017adammethodstochasticoptimization}.

\textbf{Conditional likelihood.}
\theoryrev{Under the usual regularity conditions for a conditional continuous normalizing flow~\citep{lipman2023flow}, its log-density is evaluated along the model ODE trajectory:}
\begin{equation}
  \revisioncolor
  \log p_\theta(\boldsymbol{b} \mid \boldsymbol{c})
  = \log p_0(\boldsymbol{y}_0)
  -
  \int_0^1 \operatorname{div}_{\boldsymbol{y}}
  f_\theta(\boldsymbol{y}_t,\boldsymbol{c},t)\,dt,
  \label{eq:exact-logp}
\end{equation}
\theoryrev{Here $\dot{\boldsymbol y}_t=f_\theta(\boldsymbol y_t,\boldsymbol c,t)$ is integrated backward from $\boldsymbol y_1=\boldsymbol b$. This model ODE trajectory generally differs from the sampled straight interpolation paths.}
\theoryrev{Evaluating~\eqref{eq:exact-logp} requires solving the ODE and integrating the divergence of $f_\theta$ along its trajectory. Flow-CAD scores rows through batched velocity-field evaluations without computing the density.}

\Needspace{5\baselineskip}
\subsection{Phase 2: Residual-Based Anomaly Scoring with MAD Normalization and Truncation}
\label{sec:scoring}

\begin{theoryrevision}
\textbf{Conditional denoising interpretation.}
Flow Matching regresses $\boldsymbol b-\boldsymbol z$ on state, context, and time. The induced reconstruction is $m_\theta(\boldsymbol x,\boldsymbol c,t)=\boldsymbol x+(1-t)f_\theta(\boldsymbol x,\boldsymbol c,t)$. Along the sampled path $\boldsymbol x_t$, its error relative to $\boldsymbol b$ is $(1-t)$ times the signed velocity residual. At the population optimum the reconstruction equals the posterior mean. The residual for an individual endpoint may remain nonzero. Proposition~\ref{prop:score-stability} relates field estimation error to our clipped conditional score.

\textbf{\normalcolor Raw residuals.}
Given a fitted field $f_\theta$ and an observation $(\boldsymbol c^\ast,\boldsymbol b^\ast)$, draw independent $\boldsymbol z^{(k)}\sim p_0$ for $k=1,\ldots,K_z$. For each $k$, draw fresh times $t^{(k,\ell)}\sim\mathrm{Unif}(0,1)$, $\ell=1,\ldots,K_t$. All time draws are mutually independent and independent of the Gaussian draws; the same $\boldsymbol z^{(k)}$ is reused only within its inner time loop. Set
\[
\begin{aligned}
\boldsymbol x^{(k,\ell)}
&=(1-t^{(k,\ell)})\boldsymbol z^{(k)}
   +t^{(k,\ell)}\boldsymbol b^\ast,\\
\normalcolor\boldsymbol v^{(k)}&\normalcolor=\boldsymbol b^\ast-\boldsymbol z^{(k)}.
\end{aligned}
\]
Taking the absolute value coordinatewise gives
\[
\boldsymbol r^{(k,\ell)}
=\left|f_\theta(\boldsymbol x^{(k,\ell)},\boldsymbol c^\ast,t^{(k,\ell)})
            -\boldsymbol v^{(k)}\right|.
\]
\end{theoryrevision}

\textbf{Robust per-dimension scaling via MAD.}
Behavioral dimensions can exhibit very different magnitudes and intrinsic variability. Directly aggregating raw residuals would therefore bias the anomaly score toward features with larger natural scales. To account for this, we normalize residuals using a robust, per-dimension scale estimate based on the median absolute deviation (MAD). For each behavioral dimension $q \in \{1,\dots,Q\}$, we compute
$
  \mathrm{med}_q 
  = \mathrm{median}_{i} \bigl( b_{i q} \bigr),$ and $
  \mathrm{MAD}_q
  = \mathrm{median}_{i} \bigl( |b_{i q} - \mathrm{med}_q| \bigr),
$
where $b_{iq}$ denotes the value of the $q$-th behavioral feature of the $i$-th training instance. \paperrev{We floor the scale at $\varepsilon_{\mathrm{MAD}}=10^{-6}$ and write $d_q=\max(\mathrm{MAD}_q,\varepsilon_{\mathrm{MAD}})$. The normalized residual magnitude is $\tilde r_q^{(k,\ell)}=r_q^{(k,\ell)}/d_q$.} Section~\ref{sub:exp:abaltion} ablates this scaling.

\textbf{Truncation to avoid dictator effect.}
If we simply sum the normalized residuals across behavioral dimensions, a single extremely deviant dimension could dominate the anomaly score (namely the dictator effect, also observed \paperrev{in~\citep{li2023explainable}}).
To mitigate this, we truncate each normalized residual at a fixed threshold $\tau > 0$:
$
  \hat{r}_q^{(k,\ell)}
  =
  \min\left(\tilde{r}_q^{(k,\ell)}, \tau\right),
$ with $ q = 1,\dots,Q.
$
In our experiments we use $\tau = 10$ (see Section~\ref{sub:exp:abaltion} for a sensitivity analysis on setting of this value), which can be interpreted as an upper bound on the number of ``typical MAD units'' that a single dimension is allowed to contribute. We then aggregate across dimensions as follows:
\begin{equation}
\label{eq:ResidualAgg}
  s^{(k,\ell)} = \sum_{q=1}^{Q} \hat{r}_q^{(k,\ell)}.
\end{equation}
This scalar $s^{(k,\ell)}$ is the truncated, normalized residual energy for a single pair \theoryrev{$(\boldsymbol{z}^{(k)}, t^{(k,\ell)})$}.

\textbf{Monte Carlo averaging.}
The truncated residual energy $s^{(k,\ell)}$ depends on both the sampled base noise $\boldsymbol{z}$ and the interpolation time $t$. A single realization may therefore be noisy and sensitive to stochastic
variation along the transport path. To obtain a stable anomaly score that reflects the expected mismatch between the learned conditional flow and the observed behavior, we estimate the expectation over $(\boldsymbol{z},t)$ by Monte Carlo averaging:
\begin{equation}
  s_{\mathrm{raw}}(\boldsymbol{b}^{\ast} \mid \boldsymbol{c}^{\ast})
  =
  \frac{1}{K_z K_t}
  \sum_{k=1}^{K_z}
  \sum_{\ell=1}^{K_t}
  s^{(k,\ell)}.
  \label{eq:flowcad-raw-score}
\end{equation}
\theoryrev{We use $K_z=K_t=16$, giving $256$ field evaluations in $16$ independent Gaussian-noise blocks. Proposition~\ref{prop:mc-concentration} establishes unbiasedness and quantifies the distinct effects of $K_z$ and $K_t$ on Monte Carlo variance. Section~\ref{sub:exp:abaltion} evaluates the sampling-budget trade-off empirically.}

\paperrev{\textbf{Ranking and column contributions.}}
The raw anomaly scores $s_{\mathrm{raw}}(\boldsymbol{b}\mid\boldsymbol{c})$ do not have an intrinsic probabilistic scale and may vary across datasets and sampling configurations. \paperrev{We compute detection metrics from the raw scores. Percentile ranks preserve the within-table ordering.} Because the anomaly energy in~\eqref{eq:flowcad-raw-score} is constructed from
dimension-wise residuals, Flow-CAD naturally provides per-feature contributions:
$
\mbox{$s_q(\boldsymbol b^\ast\mid\boldsymbol c^\ast)$}
=
\frac{1}{K_z K_t}
\sum_{k,\ell}
\hat{r}_q^{(k,\ell)}.
$
\paperrev{The terms sum to the raw score in~\eqref{eq:flowcad-raw-score}. Each output residual can depend on every behavioral input through the joint field. The score is decomposed by output~coordinate.}

\subsection{Theoretical Insights}
\label{sec:theory}
\begin{theoryrevision}

A row's anomaly score depends on the velocity field and the random draws used to compute it. Proposition~\ref{prop:score-stability} compares the expected scores from the fitted and optimal fields using the same MAD scales and truncation threshold. It bounds the change for an individual row and relates the average absolute change at a fixed context to the excess Flow Matching loss. With the fitted field and scoring settings fixed, Proposition~\ref{prop:mc-concentration} bounds the error from estimating the expected score with a finite number of noise and~time~draws.

With $d_q$ from Section~\ref{sec:scoring} and $0<\tau<\infty$, define
\begin{equation}
\rho(\boldsymbol u)
=\sum_{q=1}^Q\min\{|u_q|/d_q,\tau\},
\qquad
\kappa=\left(\sum_{q=1}^Q d_q^{-2}\right)^{1/2}.
\label{eq:robust-energy}
\end{equation}
For any field $f$, its expected raw score at a fixed instance is
\begin{equation}
\begin{aligned}
\mu_f(\boldsymbol b\mid\boldsymbol c)
&=\mathbb E_{\boldsymbol Z,T}
  \rho\!\left(f(\boldsymbol x_T,\boldsymbol c,T)
                 -(\boldsymbol b-\boldsymbol Z)\right),\\
\boldsymbol x_T&=(1-T)\boldsymbol Z+T\boldsymbol b,
\end{aligned}
\label{eq:population-robust-score}
\end{equation}
where $\boldsymbol Z\sim p_0$ and $T\sim\mathrm{Unif}(0,1)$ are independent. For $f=f_\theta$, this is the expected raw~score~in~\eqref{eq:flowcad-raw-score}.

\begin{proposition}[Population optimality and stability of residual scoring]
\label{prop:score-stability}
Fix a context $\boldsymbol c$. Let $\boldsymbol B\sim p(\cdot\mid\boldsymbol c)$, $\boldsymbol Z\sim p_0$, and $T\sim\mathrm{Unif}(0,1)$ be independent, with $\mathbb E\|\boldsymbol B\|_2^2<\infty$. Here $p(\cdot\mid\boldsymbol c)$ is the conditional law of the training endpoints, which can include anomalies. Set
$\boldsymbol X_t=(1-t)\boldsymbol Z+t\boldsymbol B$ and $\boldsymbol V=\boldsymbol B-\boldsymbol Z$.
Define the population risk and its optimal~regression~field~by
\begin{equation}
\begin{aligned}
\mathcal L_{\boldsymbol c}(f)
&=\mathbb E\|f(\boldsymbol X_T,\boldsymbol c,T)-\boldsymbol V\|_2^2,\\
f^\star(\boldsymbol x,\boldsymbol c,t)
&=\mathbb E[\boldsymbol V\mid
       \boldsymbol X_T=\boldsymbol x,T=t].
\end{aligned}
\label{eq:fm-optimal}
\end{equation}
Let $f_\theta$ be measurable with $\mathcal L_{\boldsymbol c}(f_\theta)<\infty$.

\emph{(i) Population risk decomposition.}
\begin{equation}
\begin{aligned}
\mathcal L_{\boldsymbol c}(f_\theta)
&=\mathcal L_{\boldsymbol c}(f^\star)
 +\mathbb E\|f_\theta(\boldsymbol X_T,\boldsymbol c,T)
                 -f^\star(\boldsymbol X_T,\boldsymbol c,T)\|_2^2,\\
\mathcal L_{\boldsymbol c}(f^\star)
&=\mathbb E\,\operatorname{tr}\operatorname{Cov}
       (\boldsymbol V\mid\boldsymbol X_T,T).
\end{aligned}
\label{eq:population-pythagorean}
\end{equation}
The expectations here include the random endpoint $\boldsymbol B$. The conditional covariance can vary with the intermediate~state~and~time.

\emph{(ii) Conditional denoising interpretation.}
For $0\le t<1$, let
$m^\star(\boldsymbol x,\boldsymbol c,t)=\mathbb E[\boldsymbol B\mid\boldsymbol X_T=\boldsymbol x,T=t]$.
At a fixed test endpoint $\boldsymbol b$, with $\boldsymbol x_t=(1-t)\boldsymbol z+t\boldsymbol b$, the posterior version of $f^\star$ satisfies
\begin{equation}
\begin{aligned}
&f^\star(\boldsymbol x_t,\boldsymbol c,t)
                    -(\boldsymbol b-\boldsymbol z)\\
&\qquad=\frac{m^\star(\boldsymbol x_t,\boldsymbol c,t)-\boldsymbol b}{1-t}.
\end{aligned}
\label{eq:posterior-denoising}
\end{equation}
Thus $\mu_{f^\star}$ is the expectation of the dimension-wise scaled and truncated denoising residual in~\eqref{eq:posterior-denoising}.

\emph{(iii) Stability of the implemented robust score.}
Define the pointwise velocity estimation error
\begin{equation}
\begin{aligned}
D_\theta(\boldsymbol b\mid\boldsymbol c)
&=\mathbb E_{\boldsymbol Z,T}
   \|f_\theta(\boldsymbol x_T,\boldsymbol c,T)
          -f^\star(\boldsymbol x_T,\boldsymbol c,T)\|_2^2.
\end{aligned}
\label{eq:pointwise-field-error}
\end{equation}
Whenever this quantity is finite,
\begin{equation}
\begin{aligned}
|\mu_{f_\theta}(\boldsymbol b\mid\boldsymbol c)
 -\mu_{f^\star}(\boldsymbol b\mid\boldsymbol c)|
&\le\kappa\sqrt{D_\theta(\boldsymbol b\mid\boldsymbol c)}.
\end{aligned}
\label{eq:pointwise-score-stability}
\end{equation}
In particular, averaging over $\boldsymbol B\sim p(\cdot\mid\boldsymbol c)$ gives
\begin{equation}
\begin{aligned}
&\mathbb E_{\boldsymbol B}
 |\mu_{f_\theta}(\boldsymbol B\mid\boldsymbol c)
  -\mu_{f^\star}(\boldsymbol B\mid\boldsymbol c)|\\
&\qquad\le\kappa
 \sqrt{\mathcal L_{\boldsymbol c}(f_\theta)
                   -\mathcal L_{\boldsymbol c}(f^\star)}.
\end{aligned}
\label{eq:population-score-stability}
\end{equation}
\end{proposition}

\textbf{From training loss to row scores.}
Part (i) splits the average training loss into the loss of the optimal field and the error from estimating it. For a fixed row, the cross term can remain, so the decomposition does not determine the row's rank. Part (ii) explains why the optimal field can still leave a residual: the posterior mean can differ from the~observed~behavior.

Part (iii) bounds how much using the fitted field changes each row's expected score. For rows $i$ and $j$ scored with the same MAD scales, let $\Delta^\star$ be the expected score of $i$ minus that of $j$ under the optimal field. The fitted field still gives $i$ a higher expected score if
\[
\Delta^\star>\kappa(\sqrt{D_i}+\sqrt{D_j}).
\]
The right-hand side adds the two bounds in~\eqref{eq:pointwise-score-stability}, with velocity estimation errors $D_i$ and $D_j$ from~\eqref{eq:pointwise-field-error}. The average training loss does not determine this gap or how it changes with training duration or network size. In the one-dimensional Gaussian example in Section~\ref{sec:proof-prop1}, even the optimal field gives higher expected scores to observations farther from~the~mean.

\begin{proposition}[Nested Monte Carlo scoring]
\label{prop:mc-concentration}
Fix $(\boldsymbol c,\boldsymbol b)$, $f_\theta$, the scales $d_q$, and $\tau$ before drawing the scoring samples. Write
$s(\boldsymbol z,t)=\rho(f_\theta(\boldsymbol x_t,\boldsymbol c,t)-(\boldsymbol b-\boldsymbol z))$,
$\mu=\mu_{f_\theta}(\boldsymbol b\mid\boldsymbol c)$, and $H=Q\tau$.
For integers $K_z,K_t\ge1$, draw $\boldsymbol Z_k$ independently from $p_0$ and draw all $T_{k\ell}$ independently from $\mathrm{Unif}(0,1)$, independently of the Gaussian draws. Define
\begin{equation}
A_k=\frac1{K_t}\sum_{\ell=1}^{K_t}s(\boldsymbol Z_k,T_{k\ell}),
\qquad
\widehat\mu=\frac1{K_z}\sum_{k=1}^{K_z}A_k.
\label{eq:nested-estimator}
\end{equation}
For the same draws in~\eqref{eq:flowcad-raw-score}, $\widehat\mu=s_{\mathrm{raw}}(\boldsymbol b\mid\boldsymbol c)$. Let
\begin{equation}
\begin{aligned}
\bar s(\boldsymbol z)&=\mathbb E_T s(\boldsymbol z,T),\\
\sigma_z^2&=\operatorname{Var}_{\boldsymbol Z}\bar s(\boldsymbol Z),\\
\sigma_t^2&=\mathbb E_{\boldsymbol Z}
          \operatorname{Var}(s(\boldsymbol Z,T)\mid\boldsymbol Z).
\end{aligned}
\label{eq:nested-variance-components}
\end{equation}
The estimator is unbiased and has exact variance
\begin{equation}
\mathbb E\widehat\mu=\mu,
\qquad
\operatorname{Var}(\widehat\mu)
 =\frac{\sigma_z^2}{K_z}+\frac{\sigma_t^2}{K_zK_t}.
\label{eq:nested-variance}
\end{equation}
Since the block means $A_k$ are independent and lie in $[0,H]$, for every $\epsilon>0$,
\begin{equation}
\Pr(|\widehat\mu-\mu|\ge\epsilon)
\le2\exp\!\left(-\frac{2K_z\epsilon^2}{H^2}\right).
\label{eq:nested-hoeffding}
\end{equation}
Equivalently, for $0<\delta<1$, with probability at least $1-\delta$,
\begin{equation}
|\widehat\mu-\mu|
\le H\sqrt{\frac{\log(2/\delta)}{2K_z}}.
\label{eq:nested-radius}
\end{equation}
A variance-sensitive bound, with $\nu_{K_t}=\sigma_z^2+\sigma_t^2/K_t$, is
\begin{equation}
\Pr(|\widehat\mu-\mu|\ge\epsilon)
\le2\exp\!\left(
 -\frac{K_z\epsilon^2}{2\nu_{K_t}+2H\epsilon/3}\right).
\label{eq:nested-bernstein}
\end{equation}
When $\nu_{K_t}=0$, the estimator equals $\mu$ almost surely.
\end{proposition}

\textbf{Choosing the number of samples.}
For a fixed row, each Gaussian draw defines one path. Equation~\eqref{eq:nested-variance} separates variation across paths from variation across sampled times on the same path. Taking more time samples ($K_t$) gives a better average along each path, but leaves the variance term $\sigma_z^2/K_z$ unchanged. Drawing more paths ($K_z$) reduces both terms, so the estimate converges to the expected score in mean square even with $K_t$ fixed.

The default $K_z=K_t=16$ uses $16$ independent paths and evaluates the field at $16$ sampled times on each, for $256$ evaluations per row. Section~\ref{sub:exp:abaltion} examines how the sampling budgets affect detection and the stability of review lists. Two rows with close expected scores can still switch places when rescored. The argument following~\eqref{eq:ranking-radius} gives a sufficient gap to preserve their order with~high~probability.

\normalcolor
\subsubsection{Proof of Proposition 1}
\label{sec:proof-prop1}

\leavevmode\revisioncolor
\textbf{Step 1: Population orthogonality.}
Let $\mathcal G=\sigma(\boldsymbol X_T,T)$ at the fixed context. Conditional expectation gives
$\mathbb E[\boldsymbol V-f^\star(\boldsymbol X_T,\boldsymbol c,T)\mid\mathcal G]=\boldsymbol0$.
Writing $\boldsymbol U=f_\theta(\boldsymbol X_T,\boldsymbol c,T)-f^\star(\boldsymbol X_T,\boldsymbol c,T)$, which~is~\mbox{$\mathcal G$-measurable,}~yields
\[
\mathbb E\langle\boldsymbol U,
          f^\star(\boldsymbol X_T,\boldsymbol c,T)-\boldsymbol V\rangle=0.
\]
All terms are integrable by the second-moment assumptions and conditional Jensen's inequality. Expanding the squared residual proves the first identity in~\eqref{eq:population-pythagorean}. The remaining conditional squared norm has expectation
\[
\mathbb E\|\boldsymbol V-\mathbb E[\boldsymbol V\mid\mathcal G]\|_2^2
=\mathbb E\,\operatorname{tr}\operatorname{Cov}(\boldsymbol V\mid\mathcal G),
\]
which proves the second identity and the optimality of $f^\star$~\citep{abu2012learning,lipman2023flow}. This orthogonality averages over the random endpoint $\boldsymbol B$; conditioning additionally on a fixed test endpoint would change the regression problem and does not justify the~same~decomposition.

\textbf{Step 2: Posterior denoising identity.}
For $t<1$, the interpolation identity implies
$\boldsymbol V=(\boldsymbol B-\boldsymbol X_t)/(1-t)$.
Applying this identity at $T$ and conditioning on $(\boldsymbol X_T,T)$ gives
\[
f^\star(\boldsymbol x,\boldsymbol c,t)
=\frac{m^\star(\boldsymbol x,\boldsymbol c,t)-\boldsymbol x}{1-t}.
\]
For a fixed $(\boldsymbol b,\boldsymbol z)$, one also has
$\boldsymbol b-\boldsymbol z=(\boldsymbol b-\boldsymbol x_t)/(1-t)$.
Subtracting proves~\eqref{eq:posterior-denoising}. The Gaussian corruption has a strictly positive density for $t<1$, so posterior versions can be used along these fixed-endpoint paths almost surely. The singular endpoint $t=1$ has zero probability under the time-sampling law and is not needed for this identity.

\textbf{Step 3: Stability of the robust energy.}
The absolute value and the clipping map are nonexpansive. Hence, for any $\boldsymbol u,\boldsymbol v\in\mathbb R^Q$,
\begin{equation}
\begin{aligned}
|\rho(\boldsymbol u)-\rho(\boldsymbol v)|
&\le\sum_{q=1}^Q\frac{|u_q-v_q|}{d_q}\\
&\le\left(\sum_{q=1}^Qd_q^{-2}\right)^{1/2}
        \|\boldsymbol u-\boldsymbol v\|_2
=\kappa\|\boldsymbol u-\boldsymbol v\|_2.
\end{aligned}
\label{eq:robust-lipschitz}
\end{equation}
At a fixed endpoint $\boldsymbol b$, apply this inequality to the two fields' residual vectors using the same $(\boldsymbol Z,T)$. The target $\boldsymbol b-\boldsymbol Z$ cancels. Taking expectations and applying~\mbox{Cauchy--Schwarz}~gives
\[
\begin{aligned}
&|\mu_{f_\theta}(\boldsymbol b\mid\boldsymbol c)
 -\mu_{f^\star}(\boldsymbol b\mid\boldsymbol c)|\\
&\quad\le\kappa\,\mathbb E_{\boldsymbol Z,T}
    \|f_\theta(\boldsymbol x_T,\boldsymbol c,T)
             -f^\star(\boldsymbol x_T,\boldsymbol c,T)\|_2\\
&\quad\le\kappa\sqrt{D_\theta(\boldsymbol b\mid\boldsymbol c)}.
\end{aligned}
\]
Finally, average over $\boldsymbol B$ and use Jensen's inequality and~Step~1:
\[
\begin{aligned}
\mathbb E_{\boldsymbol B}|\mu_{f_\theta}-\mu_{f^\star}|
&\le\kappa\sqrt{\mathbb E_{\boldsymbol B}
                         D_\theta(\boldsymbol B\mid\boldsymbol c)}\\
&=\kappa\sqrt{\mathcal L_{\boldsymbol c}(f_\theta)
                         -\mathcal L_{\boldsymbol c}(f^\star)}.
\end{aligned}
\]
This proves~\eqref{eq:population-score-stability}. The argument requires no differentiability or ODE assumptions, and applies to the ReLU field \mbox{used in our implementation}.
\qed

\textbf{A nonzero residual at the optimal field.}
For a one-dimensional reference distribution $B\sim\mathcal N(0,1)$ and independent $Z\sim\mathcal N(0,1)$, let $a(t)=t^2+(1-t)^2$. Suppressing the fixed context,
$f^\star(x,t)=(2t-1)x/a(t)$ and
\[
f^\star((1-t)z+tb,t)-(b-z)
=\frac{tz-(1-t)b}{a(t)}.
\]
Consequently, the unnormalized quadratic residual energy at a fixed $b$ is
\[
\begin{aligned}
\mathcal R^\star(b)
&=\int_0^1\frac{t^2+(1-t)^2b^2}{a(t)^2}\,dt
=\frac\pi4(1+b^2).
\end{aligned}
\]
For the implemented score, fix $0<t<1$ and write $e_t$ for the displayed residual evaluated at $z=Z$. The distribution of $|e_t|$ is that of $|G_t|$ for $G_t\sim\mathcal N(m_t,s_t^2)$, where $m_t=(1-t)|b|/a(t)$ and $s_t=t/a(t)$. For $r>0$ its CDF is
\[
\Phi((r-m_t)/s_t)+\Phi((r+m_t)/s_t)-1.
\]
Here $\Phi$ is the standard normal CDF. The displayed CDF decreases strictly with $m_t>0$, since the standard normal density at $(r+m_t)/s_t$ is smaller than at $(r-m_t)/s_t$. Using $\mathbb E\min\{|e_t|/d_1,\tau\}=\int_0^\tau\Pr(|e_t|>d_1u)\,du$ and averaging over $t$ shows that the expected implemented score strictly increases with $|b|$. It therefore orders Gaussian tail observations by increasing negative log-likelihood in this one-dimensional setting~even~at~$f^\star$.

\normalcolor
\subsubsection{Proof of Proposition 2}
\label{sec:proof-mc}

\leavevmode\revisioncolor
\textbf{Step 1: Unbiasedness under nested sampling.}
Each pair $(\boldsymbol Z_k,T_{k\ell})$ has the same marginal distribution as an independent $(\boldsymbol Z,T)$. By linearity,
\[
\mathbb E\widehat\mu
=\frac1{K_zK_t}\sum_{k,\ell}
   \mathbb E s(\boldsymbol Z_k,T_{k\ell})=\mu.
\]
This step does not require independence of all~$K_zK_t$~residual~terms.

\textbf{Step 2: Exact variance.}
Conditional on $\boldsymbol Z_k$, the $K_t$ terms within block $k$ are independent. Therefore,
\[
\begin{aligned}
\mathbb E[A_k\mid\boldsymbol Z_k]&=\bar s(\boldsymbol Z_k),\\
\operatorname{Var}(A_k\mid\boldsymbol Z_k)
&=\frac{\operatorname{Var}(s(\boldsymbol Z_k,T)\mid\boldsymbol Z_k)}{K_t}.
\end{aligned}
\]
The law of total variance gives
$\operatorname{Var}(A_k)=\sigma_z^2+\sigma_t^2/K_t=\nu_{K_t}$.
The complete random blocks $(\boldsymbol Z_k,T_{k1},\ldots,T_{kK_t})$ are independent and identically distributed, so the $A_k$ are also independent and identically distributed. Dividing their variance by $K_z$ proves~\eqref{eq:nested-variance}. In contrast, for $\ell\ne m$ within one block,
\[
\operatorname{Cov}\big(s(\boldsymbol Z_k,T_{k\ell}),
                       s(\boldsymbol Z_k,T_{km})\big)=\sigma_z^2.
\]
This covariance is generally nonzero.

\textbf{Step 3: Finite-sample concentration.}
Since $0\le s(\boldsymbol z,t)\le H$, every block average lies in $[0,H]$. Applying Hoeffding's inequality~\citep{hoeffding1963probability} to the $K_z$ independent block averages gives~\eqref{eq:nested-hoeffding}; solving its right-hand side for $\delta$ gives~\eqref{eq:nested-radius}.
For the variance-sensitive form, let $W_k=A_k-\mu$. Then
$\mathbb E W_k=0$, $|W_k|\le H$, and $\mathbb E W_k^2=\nu_{K_t}$.
For $0<\lambda<3/H$, the exponential series and
$\mathbb E|W_k|^m\le\nu_{K_t}H^{m-2}$ for $m\ge2$ yield
\[
\begin{aligned}
\log\mathbb E e^{\lambda W_k}
&\le\frac{\lambda^2\nu_{K_t}}{2(1-\lambda H/3)}.
\end{aligned}
\]
Here we used $m!\ge2\,3^{m-2}$ and $\log(1+x)\le x$. Independence and the exponential Markov inequality, with
$\lambda=\epsilon/(\nu_{K_t}+H\epsilon/3)$ when $\nu_{K_t}>0$, give
\[
\Pr(\widehat\mu-\mu\ge\epsilon)
\le\exp\!\left(-\frac{K_z\epsilon^2}
                       {2\nu_{K_t}+2H\epsilon/3}\right).
\]
Apply the same argument to $-W_k$ and add the two tails to obtain~\eqref{eq:nested-bernstein}~\citep{boucheron2013concentration}. If $\nu_{K_t}=0$, each block average is constant almost surely, which gives the stated degenerate case.
\qed

\textbf{Consequence for score ordering.}
Let $\mu_i$ and $\widehat\mu_i$ be the expected and estimated raw scores for row $i$. For $N$ fixed rows with common $Q$, $\tau$, and $K_z$, the union bound gives, with probability at least $1-\delta$ for $0<\delta<1$,
\begin{equation}
\max_{1\le i\le N}|\widehat\mu_i-\mu_i|
\le r_N:=H\sqrt{\frac{\log(2N/\delta)}{2K_z}}.
\label{eq:ranking-radius}
\end{equation}
Thus a pair with $\mu_i-\mu_j>2r_N$ retains its strict order on this event. \paperrev{At $N=20{,}000$, $K_z=16$, and $\delta=0.05$, this bound's sufficient gap is $1.30H$, exceeding the score range. We measure review-list stability under repeated scoring in Section~\ref{sec:review-list-stability}.}

\end{theoryrevision}

\section{Experiments}
\label{sec:results}
Our experiments are designed to answer the following \paperrev{research} questions: \textbf{RQ1.} \paperrev{How well does Flow-CAD rank anomalous rows?} \textbf{RQ2.} \paperrev{How does it compare with existing contextual and point detectors?} \textbf{RQ3.} \paperrev{How do behavioral features contribute to individual anomaly scores?} \textbf{RQ4.} \paperrev{How sensitive is it to scoring settings and context encoding?}

\subsection{Experimental Setup}
\label{sec:experiments}

\paperrev{This subsection describes the common experimental protocol. We introduce the datasets and their context--behavior partitions, describe how the benchmark tables and anomaly labels are constructed, explain how the baseline methods are configured, and then give the fixed Flow-CAD settings and evaluation metrics.}

\paperrev{\textbf{Datasets and feature partitions.}}
We evaluate all methods on $20$ real-world tabular datasets summarized in
Table~\ref{tab:data_summary}. \paperrev{We use the source tables released with QCAD~\citep{li2023explainable}.} We assume the partitions of features are given beforehand. \paperrev{Behavior is numerical. Context may include categorical columns.}

For example, in the \emph{Abalone} dataset, \emph{Sex, Length, Diameter, Height} are treated as contextual features and the remaining attributes as behavioral features. In ForestFires, spatial and temporal variables (\emph{X, Y, month, day}) define the context, and \paperrev{weather, fire-weather indices, and burned area form the behavior.}
\begin{table}[ht]
\centering
\caption{\paperrev{Reconstructed benchmark tables used for full-table evaluation. Column counts are reported before encoding.}}
\label{tab:data_summary}
\resizebox{\linewidth}{!}{
\begin{tabular}{lcccc}
\toprule
Dataset & \#Samples & \#Anomalies (\paperrev{Rate}) &
\paperrev{\shortstack[c]{\#Contextual\\columns}} &
\paperrev{\shortstack[c]{\#Behavioral\\columns}} \\
\midrule
Abalone       & 4177 & 100 (2.4\%) & 4 & 5 \\
\paperrev{Airfoil}       & 1503 & 70 (4.7\%) & 5 & 1 \\
BodyFat       & 252 & 20 (7.9\%) & \paperrev{12} & 2 \\
Boston        & 506 & 40 (7.9\%) & 13 & 1 \\
Concrete      & 1030 & 50 (\paperrev{4.9}\%) & 8 & 1 \\
ElNino        & 20000 & 200 (1.0\%) & 6 & 5 \\
Energy        & 768 & 50 (6.5\%) & 8 & 2 \\
FishWeight    & \paperrev{159} & 15 (\paperrev{9.4}\%) & 6 & 1 \\
ForestFires   & 517 & 50 (9.7\%) & 4 & 9 \\
GasEmission   & 7384 & 100 (1.4\%) & 8 & 3 \\
HeartFailure  & 299 & 30 (10.0\%) & 6 & 6 \\
Hepatitis     & \paperrev{589} & 30 (\paperrev{5.1}\%) & 3 & 10 \\
IndianLiver   & 579 & 30 (5.2\%) & 3 & 8 \\
Maintenance   & 11934 & 100 (0.8\%) & 15 & 3 \\
Parkinson     & 5875 & 100 (1.7\%) & 20 & 2 \\
PowerPlant    & 9568 & 100 (1.0\%) & 4 & 1 \\
QSRanking     & 475 & 40 (8.4\%) & 7 & 6 \\
SynMachine    & 557 & 50 (\paperrev{9.0}\%) & 4 & 1 \\
Toxicity      & 908 & 50 (5.5\%) & 6 & 1 \\
Yacht         & 308 & 30 (9.7\%) & 6 & 1 \\
\bottomrule
\end{tabular}
}
\end{table}

\textbf{Benchmark construction.} Following the QCAD benchmark~\citep{li2023explainable}, \paperrev{we remove incomplete rows, cap each input at $20{,}000$ rows by seeded subsampling, and min--max scale the numerical columns designated by the benchmark. We sample the number of rows listed in Table~\ref{tab:data_summary} uniformly without replacement, excluding the first row as in the released generator. Each selected row receives an independent signed shift in every behavioral column, with magnitude drawn uniformly from $[0.1,0.5]$ and the sign chosen uniformly. The selected rows define the benchmark anomaly labels.}

\paperrev{We use seven data seeds ($43$--$49$) for each dataset. All methods are evaluated on the same $140$ benchmark tables, with runs paired by dataset and data seed.}

\textbf{Baselines.} We compare Flow-CAD against 13 representative baselines, including four contextual anomaly detectors (QCAD~\citep{li2023explainable},
LoPAD~\citep{lu2020lopad},
ROCOD~\citep{liang2016robust},
CAD~\citep{song2007conditional}), and nine point anomaly detectors (%
Isolation Forest~\citep{liu2008isolation},
LOF~\citep{breunig2000lof},
$k$-NN~\citep{angiulli2002fast},
SOD~\citep{kriegel2009outlier}, 
HBOS~\citep{goldstein2012histogram}, LUNAR~\citep{goodge2022lunar}, DTE-Categorical, DTE-NonParametric~\citep{livernoche2024on}, and TCCM~\citep{li2025scalable}).

\paperrev{We keep each baseline's settings fixed across runs on a given dataset. We reimplemented QCAD using the neighborhood sizes specified in its benchmark. LoPAD uses the Markov blankets from the QCAD release and fits CART models on $20$ subsamples. ROCOD uses the benchmark distance cutoff and relies on its global predictor for rows with no neighbors. CAD fits a joint Gaussian mixture with five components and scores rows by their conditional density. We use the authors' released code for DTE and TCCM. For the remaining methods, we use the settings from PyOD 2.0.3~\citep{zhao2019pyod}, including LUNAR's built-in validation with~synthetic~negatives.}

\paperrev{\textbf{Flow-CAD configuration.}} Flow-CAD one-hot encodes the designated categorical context columns. \paperrev{The resulting context vector is held fixed along the interpolation path, while interpolation and feature attribution are defined only for the numerical behavioral features.}
\paperrev{We use the fixed settings in Table~\ref{tab:flowcad-hparams} for every run. Each model is trained on a complete benchmark table for $1{,}000$ epochs, and its final checkpoint is used to score all rows in that table. The seven runs pair data seeds $43$--$49$ with training seeds $202$--$208$ in ascending order. Each run uses a single CPU thread.}

\begin{table}[b]
    \centering
    \caption{\paperrev{Flow-CAD configuration fixed across benchmark datasets and runs.}}
    \label{tab:flowcad-hparams}
    \scriptsize
    \begin{tabular}{ll}
        \toprule
        Component & Fixed value \\
        \midrule
        Context encoder & \paperrev{$2$ layers of width $32$} \\
        Velocity network & \paperrev{$3$ hidden layers of width $64$, ReLU} \\
        Optimizer & Adam, learning rate $0.001$ \\
        \paperrev{Batch size} & \paperrev{$128$} \\
        \paperrev{Training epochs} & \paperrev{$1{,}000$} \\
        Score aggregation & $L_1$ \\
        \paperrev{Residual truncation} & $\tau=10$ \\
        Monte Carlo budget & $K_z=K_t=16$ \\
        \bottomrule
    \end{tabular}
\end{table}

\textbf{Evaluation metrics.} Anomaly labels are reserved for evaluation. For the main same-table benchmark, we report \paperrev{the area under the receiver operating characteristic curve (ROC AUC), the trapezoidal area under the precision--recall curve (PR AUC), and precision at $n$ (P@$n$)}, with $n$ equal to the number of labeled anomalies. We report mean and \paperrev{sample} standard deviation across runs and average the per-dataset ranks over the $20$ datasets.

\Needspace{4\baselineskip}
\subsection{\texorpdfstring{\paperrev{Detection Quality Across Benchmark Tables} (RQ1)}{Detection Quality Across Benchmark Tables (RQ1)}}
\label{sec:detailed-main-results}

\paperrev{Tables~\ref{tab:main_results_pr}--\ref{tab:main_results_pn} report results on $20$ datasets. We analyze the overall ranks and detection when anomalies are rare, then leave out one dataset at a time to test rank stability.}

\paperrev{\textbf{1) Ranks across datasets.} Among the $14$ methods, Flow-CAD has the best average rank for PR AUC ($1.900$), ROC AUC ($2.050$), and P@$n$ ($1.825$). These metrics examine different parts of detection: ROC AUC summarizes the full score ordering, PR AUC focuses on correctly retrieving anomalies, and P@$n$ measures precision among the top $n$ rows. Thus, Flow-CAD performs well both across the full ranking and among the rows inspected first.}

\paperrev{\textbf{2) Detection with rare anomalies.} Five datasets in Table~\ref{tab:data_summary} have anomaly rates below $2\%$. Across their $15$ dataset--metric comparisons, Flow-CAD ranks among the top three methods in $13$. It ranks first on all three metrics for Parkinson, and first for PR AUC and P@$n$ on PowerPlant. These results show that its strong overall ranking is retained when only a small fraction of rows is anomalous.}

\paperrev{\textbf{3) Influence of individual datasets.} Because an average rank can be driven by one especially favorable dataset, we recalculate it with one dataset left out at a time. Each calculation therefore averages the ranks over the other $19$ datasets. Flow-CAD remains first for PR AUC and P@$n$ in all $20$ cases. For ROC AUC, it ranks first or ties for first in $16$ cases and ranks second in the other~four. Thus, the conclusions for PR AUC and P@$n$ do not depend on any single table; the ROC AUC comparison is closer, but Flow-CAD never falls below second.}


\begin{table}[tbp]
\centering
\caption{\paperrev{PR AUC on $20$ datasets. Entries give mean $\pm$ standard deviation (rank). Ranks use unrounded means and average numerical ties ($10^{-12}$ tolerance). Equal displayed means can have different ranks. Bold and underlining mark the top two distinct ranks. Each method has 140 paired cases, with seven benchmark runs per dataset.}}
\label{tab:main_results_pr}
\scriptsize
\setlength{\tabcolsep}{4pt}
\resizebox{\columnwidth}{!}{
\begin{tabular}{lccccc}
\toprule
Dataset & Flow-CAD (ours) & QCAD & LoPAD & ROCOD & CAD \\
\midrule
Abalone & \textbf{\paperrev{1.00}}$\pm$0.00 (1) & 0.96$\pm$0.01 (\paperrev{3}) & \paperrev{0.01}$\pm$0.00 (14) & \paperrev{0.55$\pm$0.03 (10)} & \underline{\paperrev{0.98}}$\pm$\paperrev{0.01} (\paperrev{2})\\
Airfoil & \paperrev{0.44$\pm$0.10 (4)} & \paperrev{\textbf{0.68}$\pm$0.05 (1)} & 0.05$\pm$\paperrev{0.00} (\paperrev{14}) & \paperrev{0.42$\pm$0.04 (5)} & \paperrev{\underline{0.49}$\pm$0.04 (2)}\\
BodyFat & \paperrev{\textbf{0.84}$\pm$0.08 (1)} & \paperrev{0.63$\pm$0.08 (4)} & \paperrev{0.07$\pm$0.02 (14)} & \paperrev{0.79$\pm$0.05 (3)} & \paperrev{\underline{0.81}$\pm$0.12 (2)}\\
Boston & \paperrev{\underline{0.60}$\pm$0.10 (2)} & \paperrev{\textbf{0.63}$\pm$0.10 (1)} & \paperrev{0.08$\pm$0.00 (13)} & 0.30$\pm$0.06 (4) & \paperrev{0.32$\pm$0.09 (3)}\\
Concrete & \paperrev{\textbf{0.67}$\pm$0.14 (1)} & \paperrev{\underline{0.60}$\pm$0.09 (2)} & \paperrev{0.09$\pm$0.02 (9)} & \paperrev{0.31$\pm$0.06 (5)} & \paperrev{0.30$\pm$0.06 (6)}\\
ElNino & \paperrev{0.81$\pm$0.04 (3)} & \textbf{\paperrev{0.99}}$\pm$\paperrev{0.00} (1) & 0.11$\pm$0.01 (\paperrev{8}) & \paperrev{0.39$\pm$0.05 (7)} & \underline{\paperrev{0.96}}$\pm$\paperrev{0.01} (\paperrev{2})\\
Energy & \textbf{\paperrev{0.94}}$\pm$\paperrev{0.03} (1) & \paperrev{\underline{0.93}$\pm$0.02 (2)} & 0.05$\pm$\paperrev{0.00} (14) & \paperrev{0.43$\pm$0.07 (7)} & \paperrev{0.91$\pm$0.03 (3)}\\
FishWeight & \paperrev{\textbf{0.90}$\pm$0.05 (1)} & \paperrev{\underline{0.85}$\pm$0.10 (2)} & \paperrev{0.13$\pm$0.06 (8)} & \paperrev{0.44$\pm$0.10 (4)} & \paperrev{0.72$\pm$0.09 (3)}\\
ForestFires & \paperrev{0.97}$\pm$\paperrev{0.02} (4) & \underline{0.99}$\pm$\paperrev{0.01} (2) & 0.10$\pm$0.01 (14) & \paperrev{0.49$\pm$0.03 (10)} & \paperrev{0.88$\pm$0.03 (6)}\\
GasEmission & \underline{0.98}$\pm$0.01 (\paperrev{2}) & \textbf{\paperrev{0.98}}$\pm$\paperrev{0.00} (\paperrev{1}) & 0.01$\pm$0.00 (\paperrev{14}) & \paperrev{0.87$\pm$0.02 (3)} & \paperrev{0.85$\pm$0.02 (4)}\\
HeartFailure & \underline{\paperrev{0.97}}$\pm$\paperrev{0.01} (\paperrev{2}) & \paperrev{0.91$\pm$0.03 (3)} & 0.05$\pm$0.00 (14) & \paperrev{0.54$\pm$0.03 (6)} & \paperrev{0.77$\pm$0.09 (5)}\\
Hepatitis & \underline{0.99}$\pm$0.01 (\paperrev{2}) & 0.98$\pm$0.01 (3) & 0.05$\pm$0.01 (14) & 0.55$\pm$0.03 (\paperrev{8}) & \paperrev{0.84$\pm$0.08 (5)}\\
IndianLiver & \underline{\paperrev{0.97}}$\pm$0.01 (2) & \paperrev{0.90$\pm$0.04 (4)} & 0.05$\pm$0.00 (14) & \paperrev{0.50$\pm$0.06 (8)} & \paperrev{0.79$\pm$0.12 (6)}\\
Maintenance & \paperrev{0.76$\pm$0.08 (4)} & \paperrev{\textbf{0.95}$\pm$0.02 (1)} & \paperrev{0.01$\pm$0.00 (12)} & 0.50$\pm$0.00 (\paperrev{5}) & \paperrev{0.87$\pm$0.04 (3)}\\
Parkinson & \textbf{\paperrev{0.98}}$\pm$0.01 (1) & \underline{\paperrev{0.92}}$\pm$\paperrev{0.02} (2) & \paperrev{0.02$\pm$0.00 (14)} & \paperrev{0.60$\pm$0.06 (4)} & \paperrev{0.75$\pm$0.04 (3)}\\
PowerPlant & \paperrev{\textbf{0.66}$\pm$0.04 (1)} & \paperrev{0.46$\pm$0.04 (4)} & 0.02$\pm$0.00 (\paperrev{10}) & \paperrev{\underline{0.66}$\pm$0.05 (2)} & \paperrev{0.56$\pm$0.07 (3)}\\
QSRanking & \paperrev{0.74$\pm$0.08 (3)} & \paperrev{\textbf{0.78}$\pm$0.05 (1)} & 0.06$\pm$0.00 (14) & \paperrev{0.56$\pm$0.06 (4)} & \paperrev{\underline{0.77}$\pm$0.07 (2)}\\
SynMachine & \textbf{1.00}$\pm$0.00 (1) & \paperrev{0.97$\pm$0.02 (3)} & \paperrev{0.35$\pm$0.03 (13)} & \paperrev{0.80$\pm$0.07 (7)} & \paperrev{0.54$\pm$0.12 (11)}\\
Toxicity & \paperrev{\textbf{0.61}$\pm$0.06 (1)} & \paperrev{0.44$\pm$0.09 (4)} & \paperrev{0.11$\pm$0.01 (8)} & 0.46$\pm$0.06 (3) & \paperrev{\underline{0.46}$\pm$0.06 (2)}\\
Yacht & \paperrev{\textbf{0.92}$\pm$0.06 (1)} & \underline{\paperrev{0.89}}$\pm$0.04 (2) & 0.53$\pm$\paperrev{0.01} (\paperrev{3}) & \paperrev{0.29$\pm$0.06 (6)} & \paperrev{0.25$\pm$0.07 (9)}\\
\midrule
\textbf{Avg Rank} & \textbf{\paperrev{1.900}} & \paperrev{\underline{2.300}} & \paperrev{11.900} & \paperrev{5.550} & \paperrev{4.100}\\
\midrule
Dataset & IForest & LOF & $k$-NN & SOD & HBOS \\
\midrule
Abalone & \paperrev{0.41$\pm$0.07 (11)} & 0.92$\pm$0.01 (\paperrev{6}) & 0.94$\pm$0.00 (\paperrev{5}) & \paperrev{0.75$\pm$0.01 (9)} & \paperrev{0.20$\pm$0.04 (13)}\\
Airfoil & 0.10$\pm$0.02 (\paperrev{8}) & 0.05$\pm$0.01 (\paperrev{11}) & 0.05$\pm$0.01 (12) & 0.14$\pm$0.02 (7) & 0.09$\pm$0.02 (\paperrev{9})\\
BodyFat & \paperrev{0.17$\pm$0.04 (7)} & \paperrev{0.09$\pm$0.02 (12)} & \paperrev{0.09$\pm$0.02 (13)} & 0.10$\pm$0.04 (\paperrev{8}) & \paperrev{0.18$\pm$0.03 (6)}\\
Boston & 0.11$\pm$0.02 (\paperrev{8}) & \paperrev{0.08$\pm$0.01 (11)} & \paperrev{0.08$\pm$0.01 (12)} & 0.08$\pm$0.01 (\paperrev{9}) & \paperrev{0.13$\pm$0.02 (6)}\\
Concrete & 0.08$\pm$0.02 (\paperrev{10}) & \paperrev{0.06$\pm$0.01 (12)} & \paperrev{0.06$\pm$0.02 (11)} & \paperrev{0.06$\pm$0.01 (13)} & \paperrev{0.26$\pm$0.06 (7)}\\
ElNino & \paperrev{0.55$\pm$0.10 (5)} & \paperrev{0.01$\pm$0.00 (14)} & \paperrev{0.01$\pm$0.00 (13)} & 0.04$\pm$0.00 (\paperrev{9}) & \paperrev{0.77$\pm$0.03 (4)}\\
Energy & \paperrev{0.39$\pm$0.03 (8)} & \paperrev{0.11$\pm$0.02 (12)} & 0.37$\pm$0.02 (\paperrev{9}) & \paperrev{0.89$\pm$0.05 (4)} & \paperrev{0.32$\pm$0.05 (11)}\\
FishWeight & \paperrev{0.22$\pm$0.05 (6)} & \paperrev{0.11$\pm$0.04 (14)} & \paperrev{0.11$\pm$0.04 (12)} & \paperrev{0.12$\pm$0.03 (11)} & \paperrev{0.20$\pm$0.07 (7)}\\
ForestFires & 0.97$\pm$0.02 (3) & 0.17$\pm$0.02 (\paperrev{13}) & 0.18$\pm$0.02 (\paperrev{12}) & \paperrev{0.57$\pm$0.04 (9)} & \textbf{1.00}$\pm$0.00 (1)\\
GasEmission & \paperrev{0.10$\pm$0.02 (7)} & \paperrev{0.01$\pm$0.00 (13)} & 0.01$\pm$0.00 (12) & \paperrev{0.03$\pm$0.00 (8)} & \paperrev{0.25$\pm$0.05 (6)}\\
HeartFailure & \paperrev{0.86$\pm$0.02 (4)} & \paperrev{0.13$\pm$0.03 (12)} & \paperrev{0.17$\pm$0.05 (11)} & \paperrev{0.37$\pm$0.07 (9)} & \paperrev{\textbf{0.98}$\pm$0.01 (1)}\\
Hepatitis & \paperrev{0.93}$\pm$\paperrev{0.04} (4) & 0.17$\pm$0.02 (12) & \paperrev{0.15$\pm$0.02 (13)} & 0.35$\pm$0.01 (\paperrev{10}) & \textbf{0.99}$\pm$0.01 (1)\\
IndianLiver & \paperrev{0.94$\pm$0.02 (3)} & 0.11$\pm$0.02 (12) & 0.17$\pm$0.04 (\paperrev{11}) & \paperrev{0.45$\pm$0.05 (9)} & \textbf{0.99}$\pm$0.01 (1)\\
Maintenance & 0.02$\pm$\paperrev{0.00} (\paperrev{10}) & 0.01$\pm$0.00 (\paperrev{14}) & \paperrev{0.01$\pm$0.00 (13)} & 0.02$\pm$0.00 (\paperrev{9}) & \paperrev{0.23$\pm$0.08 (6)}\\
Parkinson & 0.02$\pm$0.00 (10) & \paperrev{0.02$\pm$0.00 (13)} & \paperrev{0.02$\pm$0.00 (12)} & 0.04$\pm$0.00 (\paperrev{8}) & 0.02$\pm$0.00 (\paperrev{9})\\
PowerPlant & \paperrev{0.05$\pm$0.01 (9)} & \paperrev{0.01$\pm$0.00 (14)} & \paperrev{0.01$\pm$0.00 (12)} & 0.01$\pm$0.00 (\paperrev{11}) & \paperrev{0.23$\pm$0.03 (7)}\\
QSRanking & \paperrev{0.22$\pm$0.07 (7)} & \paperrev{0.10$\pm$0.03 (10)} & 0.09$\pm$0.02 (\paperrev{12}) & 0.09$\pm$0.01 (\paperrev{13}) & \paperrev{0.44$\pm$0.07 (5)}\\
SynMachine & \paperrev{0.36$\pm$0.06 (12)} & \paperrev{0.80$\pm$0.04 (8)} & 0.91$\pm$\paperrev{0.02} (4) & \paperrev{0.72$\pm$0.06 (9)} & \paperrev{0.26$\pm$0.04 (14)}\\
Toxicity & \paperrev{0.09$\pm$0.01 (10)} & 0.08$\pm$0.02 (\paperrev{11}) & 0.07$\pm$0.01 (14) & 0.12$\pm$0.02 (\paperrev{7}) & 0.10$\pm$0.01 (\paperrev{9})\\
Yacht & \paperrev{0.28$\pm$0.07 (7)} & \paperrev{0.19$\pm$0.04 (13)} & \paperrev{0.25$\pm$0.03 (10)} & \paperrev{0.25$\pm$0.05 (8)} & \paperrev{0.31$\pm$0.09 (4)}\\
\midrule
\textbf{Avg Rank} & \paperrev{7.450} & \paperrev{11.850} & \paperrev{11.150} & \paperrev{9.000} & \paperrev{6.350}\\
\midrule
Dataset & LUNAR & DTE-C & \multicolumn{2}{c}{DTE-NP} & TCCM \\
\midrule
Abalone & \paperrev{0.92$\pm$0.03 (7)} & \paperrev{0.92$\pm$0.01 (8)} & \multicolumn{2}{c}{\paperrev{0.94$\pm$0.00 (4)}} & \paperrev{0.24$\pm$0.11 (12)}\\
Airfoil & \paperrev{0.29}$\pm$\paperrev{0.06} (6) & \paperrev{0.46$\pm$0.18 (3)} & \multicolumn{2}{c}{\paperrev{0.06}$\pm$\paperrev{0.01} (\paperrev{10})} & \paperrev{0.05}$\pm$\paperrev{0.01} (\paperrev{13})\\
BodyFat & \paperrev{0.31$\pm$0.08 (5)} & \paperrev{0.09$\pm$0.05 (11)} & \multicolumn{2}{c}{\paperrev{0.10}$\pm$\paperrev{0.03} (\paperrev{9})} & \paperrev{0.10$\pm$0.02 (10)}\\
Boston & \paperrev{0.13$\pm$0.02 (7)} & \paperrev{0.07$\pm$0.01 (14)} & \multicolumn{2}{c}{\paperrev{0.16$\pm$0.03 (5)}} & \paperrev{0.08}$\pm$0.02 (\paperrev{10})\\
Concrete & \paperrev{0.12}$\pm$\paperrev{0.06} (8) & \paperrev{0.52}$\pm$0.00 (3) & \multicolumn{2}{c}{\paperrev{0.48$\pm$0.03 (4)}} & \paperrev{0.05}$\pm$0.01 (\paperrev{14})\\
ElNino & \paperrev{0.02$\pm$0.01 (10)} & \paperrev{0.51$\pm$0.00 (6)} & \multicolumn{2}{c}{\paperrev{0.01}$\pm$0.00 (\paperrev{11})} & \paperrev{0.01}$\pm$0.00 (\paperrev{12})\\
Energy & \paperrev{0.44$\pm$0.08 (6)} & \paperrev{0.53}$\pm$0.00 (\paperrev{5}) & \multicolumn{2}{c}{\paperrev{0.34$\pm$0.02 (10)}} & \paperrev{0.09$\pm$0.03 (13)}\\
FishWeight & \paperrev{0.31$\pm$0.07 (5)} & \paperrev{0.11$\pm$0.04 (13)} & \multicolumn{2}{c}{\paperrev{0.12$\pm$0.05 (10)}} & \paperrev{0.13$\pm$0.05 (9)}\\
ForestFires & \paperrev{0.78$\pm$0.16 (8)} & \paperrev{0.78$\pm$0.05 (7)} & \multicolumn{2}{c}{\paperrev{0.21}$\pm$\paperrev{0.02} (\paperrev{11})} & \paperrev{0.90$\pm$0.02 (5)}\\
GasEmission & \paperrev{0.41$\pm$0.15 (5)} & \paperrev{0.01}$\pm$\paperrev{0.00} (\paperrev{11}) & \multicolumn{2}{c}{\paperrev{0.01$\pm$0.00 (10)}} & \paperrev{0.03$\pm$0.00 (9)}\\
HeartFailure & \paperrev{0.41$\pm$0.04 (7)} & \paperrev{0.12}$\pm$\paperrev{0.02} (\paperrev{13}) & \multicolumn{2}{c}{\paperrev{0.20$\pm$0.06 (10)}} & \paperrev{0.40$\pm$0.11 (8)}\\
Hepatitis & \paperrev{0.53$\pm$0.09 (9)} & \paperrev{0.84$\pm$0.04 (6)} & \multicolumn{2}{c}{\paperrev{0.19$\pm$0.02 (11)}} & \paperrev{0.73$\pm$0.20 (7)}\\
IndianLiver & \paperrev{0.85$\pm$0.02 (5)} & \paperrev{0.08$\pm$0.03 (13)} & \multicolumn{2}{c}{\paperrev{0.21}$\pm$\paperrev{0.04} (\paperrev{10})} & \paperrev{0.59$\pm$0.13 (7)}\\
Maintenance & \paperrev{\underline{0.94}$\pm$0.08 (2)} & \paperrev{0.15$\pm$0.18 (7)} & \multicolumn{2}{c}{\paperrev{0.07$\pm$0.01 (8)}} & \paperrev{0.01}$\pm$0.00 (\paperrev{11})\\
Parkinson & \paperrev{0.14$\pm$0.02 (6)} & \paperrev{0.51}$\pm$\paperrev{0.00} (5) & \multicolumn{2}{c}{\paperrev{0.02$\pm$0.00 (11)}} & \paperrev{0.12$\pm$0.07 (7)}\\
PowerPlant & \paperrev{0.43$\pm$0.16 (6)} & \paperrev{0.43$\pm$0.19 (5)} & \multicolumn{2}{c}{\paperrev{0.01$\pm$0.00 (13)}} & \paperrev{0.15$\pm$0.11 (8)}\\
QSRanking & \paperrev{0.17$\pm$0.03 (8)} & \paperrev{0.10$\pm$0.02 (11)} & \multicolumn{2}{c}{\paperrev{0.43}$\pm$\paperrev{0.03} (\paperrev{6})} & \paperrev{0.10}$\pm$\paperrev{0.02} (9)\\
SynMachine & \paperrev{0.85$\pm$0.04 (6)} & \paperrev{0.62$\pm$0.04 (10)} & \multicolumn{2}{c}{\underline{\paperrev{0.98}}$\pm$\paperrev{0.01} (2)} & \paperrev{0.90$\pm$0.02 (5)}\\
Toxicity & \paperrev{0.15$\pm$0.04 (5)} & \paperrev{0.14}$\pm$0.01 (\paperrev{6}) & \multicolumn{2}{c}{\paperrev{0.07$\pm$0.01 (12)}} & \paperrev{0.07$\pm$0.02 (13)}\\
Yacht & \paperrev{0.23$\pm$0.06 (11)} & \paperrev{0.22$\pm$0.06 (12)} & \multicolumn{2}{c}{\paperrev{0.30$\pm$0.05 (5)}} & \paperrev{0.19$\pm$0.04 (14)}\\
\midrule
\textbf{Avg Rank} & \paperrev{6.600} & \paperrev{8.450} & \multicolumn{2}{c}{\paperrev{8.600}} & \paperrev{9.800}\\
\bottomrule

\end{tabular}
}
\end{table}

\begin{table}[tbp]
\centering
\caption{\paperrev{ROC AUC on $20$ datasets. Notation follows Table~\ref{tab:main_results_pr}.}}
\label{tab:main_results_roc}
\scriptsize
\setlength{\tabcolsep}{4pt}
\resizebox{\columnwidth}{!}{
\begin{tabular}{lccccc}
\toprule
Dataset & Flow-CAD (ours) & QCAD & LoPAD & ROCOD & CAD \\
\midrule
Abalone & \textbf{1.00}$\pm$0.00 (1) & \paperrev{1.00$\pm$0.00 (4)} & \paperrev{0.29$\pm$0.03 (14)} & 0.98$\pm$0.00 (\paperrev{9}) & \paperrev{0.99}$\pm$\paperrev{0.01} (\paperrev{8})\\
Airfoil & \paperrev{0.81$\pm$0.05 (4)} & \textbf{\paperrev{0.92}}$\pm$\paperrev{0.02} (1) & \paperrev{0.50$\pm$0.03 (13)} & 0.79$\pm$0.02 (\paperrev{5}) & \paperrev{0.82}$\pm$\paperrev{0.02} (\paperrev{3})\\
BodyFat & \underline{\paperrev{0.96}}$\pm$\paperrev{0.03} (2) & \paperrev{0.93}$\pm$\paperrev{0.04} (4) & \paperrev{0.49$\pm$0.03 (14)} & \paperrev{0.94$\pm$0.03 (3)} & \textbf{\paperrev{0.98}}$\pm$\paperrev{0.01} (1)\\
Boston & \paperrev{\underline{0.89}$\pm$0.04 (2)} & \paperrev{\textbf{0.93}$\pm$0.02 (1)} & \paperrev{0.47$\pm$0.02 (13)} & \paperrev{0.82$\pm$0.04 (3)} & \paperrev{0.78$\pm$0.05 (4)}\\
Concrete & \paperrev{\underline{0.91}$\pm$0.06 (2)} & \paperrev{\textbf{0.93}$\pm$0.03 (1)} & \paperrev{0.63$\pm$0.03 (7)} & \paperrev{0.77$\pm$0.03 (3)} & \paperrev{0.74}$\pm$0.04 (\paperrev{4})\\
ElNino & \paperrev{0.99$\pm$0.00 (3)} & \textbf{1.00}$\pm$0.00 (1) & \paperrev{0.91$\pm$0.01 (7)} & \paperrev{0.98$\pm$0.00 (5)} & \underline{\paperrev{1.00}}$\pm$\paperrev{0.00} (\paperrev{2})\\
Energy & \underline{0.99}$\pm$0.01 (2) & \textbf{0.99}$\pm$0.00 (1) & \paperrev{0.38$\pm$0.03 (14)} & \paperrev{0.82$\pm$0.04 (9)} & \paperrev{0.98}$\pm$\paperrev{0.01} (\paperrev{3})\\
FishWeight & \paperrev{\textbf{0.97}$\pm$0.03 (1)} & \paperrev{\underline{0.96}$\pm$0.03 (2)} & \paperrev{0.52$\pm$0.05 (12)} & \paperrev{0.80$\pm$0.07 (5)} & \paperrev{0.91$\pm$0.05 (3)}\\
ForestFires & \paperrev{1.00}$\pm$\paperrev{0.00} (\paperrev{4}) & \underline{1.00}$\pm$0.00 (\paperrev{2}) & \paperrev{0.51}$\pm$0.04 (14) & \paperrev{0.87$\pm$0.02 (10)} & \paperrev{0.99$\pm$0.00 (6)}\\
GasEmission & \underline{1.00}$\pm$0.00 (\paperrev{2}) & \textbf{1.00}$\pm$0.00 (1) & \paperrev{0.46$\pm$0.01 (14)} & 1.00$\pm$0.00 (\paperrev{4}) & \paperrev{1.00}$\pm$\paperrev{0.00} (\paperrev{3})\\
HeartFailure & \underline{\paperrev{1.00}}$\pm$\paperrev{0.00} (\paperrev{2}) & \paperrev{0.99$\pm$0.01 (3)} & \paperrev{0.11$\pm$0.03 (14)} & \paperrev{0.52$\pm$0.05 (13)} & 0.98$\pm$0.01 (\paperrev{5})\\
Hepatitis & \underline{1.00}$\pm$0.00 (\paperrev{2}) & 1.00$\pm$0.00 (\paperrev{3}) & \paperrev{0.50$\pm$0.07 (14)} & 0.98$\pm$0.00 (\paperrev{8}) & \paperrev{0.99$\pm$0.01 (6)}\\
IndianLiver & \underline{1.00}$\pm$\paperrev{0.00} (\paperrev{2}) & \paperrev{0.99$\pm$0.00 (4)} & \paperrev{0.48$\pm$0.06 (14)} & 0.97$\pm$0.01 (\paperrev{7}) & \paperrev{0.98$\pm$0.03 (6)}\\
Maintenance & \paperrev{0.97$\pm$0.02 (3)} & \underline{1.00}$\pm$0.00 (\paperrev{2}) & \paperrev{0.51$\pm$0.05 (11)} & \paperrev{0.50$\pm$0.00 (13)} & \paperrev{0.97}$\pm$\paperrev{0.01} (\paperrev{4})\\
Parkinson & \textbf{1.00}$\pm$0.00 (1) & \underline{0.99}$\pm$\paperrev{0.01} (2) & \paperrev{0.42$\pm$0.02 (14)} & \paperrev{0.95$\pm$0.02 (3)} & \paperrev{0.92$\pm$0.03 (5)}\\
PowerPlant & \paperrev{\underline{0.98}$\pm$0.00 (2)} & \textbf{0.98}$\pm$0.01 (\paperrev{1}) & \paperrev{0.46}$\pm$0.01 (14) & 0.97$\pm$0.01 (3) & \paperrev{0.91}$\pm$\paperrev{0.02} (\paperrev{4})\\
QSRanking & \paperrev{0.94}$\pm$\paperrev{0.03} (3) & \textbf{0.96}$\pm$0.01 (1) & \paperrev{0.37$\pm$0.03 (14)} & \paperrev{0.88$\pm$0.03 (4)} & \underline{\paperrev{0.95}}$\pm$\paperrev{0.02} (\paperrev{2})\\
SynMachine & \textbf{1.00}$\pm$0.00 (1) & 0.98$\pm$0.01 (4) & \paperrev{0.79$\pm$0.04 (12)} & \paperrev{0.90$\pm$0.03 (8)} & \paperrev{0.75}$\pm$\paperrev{0.09} (13)\\
Toxicity & \paperrev{\textbf{0.90}$\pm$0.02 (1)} & \paperrev{\underline{0.88}$\pm$0.03 (2)} & \paperrev{0.58$\pm$0.04 (13)} & \paperrev{0.85$\pm$0.04 (3)} & \paperrev{0.84}$\pm$\paperrev{0.04} (\paperrev{4})\\
Yacht & \paperrev{\textbf{0.99}$\pm$0.01 (1)} & \paperrev{\underline{0.98}$\pm$0.02 (2)} & 0.96$\pm$0.00 (3) & \paperrev{0.77$\pm$0.03 (7)} & \paperrev{0.73$\pm$0.06 (10)}\\
\midrule
\textbf{Avg Rank} & \paperrev{\textbf{2.050}} & \paperrev{\underline{2.100}} & \paperrev{12.250} & \paperrev{6.250} & \paperrev{4.800}\\
\midrule
Dataset & IForest & LOF & $k$-NN & SOD & HBOS \\
\midrule
Abalone & 0.98$\pm$0.00 (\paperrev{10}) & \paperrev{1.00$\pm$0.00 (5)} & 1.00$\pm$0.00 (\paperrev{3}) & 0.96$\pm$0.01 (\paperrev{11}) & \paperrev{0.92$\pm$0.01 (12)}\\
Airfoil & \paperrev{0.71}$\pm$\paperrev{0.04} (7) & \paperrev{0.52$\pm$0.04 (10)} & \paperrev{0.51$\pm$0.03 (11)} & \paperrev{0.78$\pm$0.03 (6)} & \paperrev{0.65$\pm$0.02 (8)}\\
BodyFat & \paperrev{0.75$\pm$0.05 (7)} & \paperrev{0.52$\pm$0.06 (10)} & \paperrev{0.52$\pm$0.05 (11)} & \paperrev{0.54$\pm$0.06 (8)} & \paperrev{0.76$\pm$0.05 (6)}\\
Boston & \paperrev{0.59}$\pm$0.04 (\paperrev{7}) & \paperrev{0.50$\pm$0.04 (9)} & \paperrev{0.48$\pm$0.04 (12)} & \paperrev{0.50$\pm$0.04 (8)} & \paperrev{0.67$\pm$0.04 (6)}\\
Concrete & \paperrev{0.61$\pm$0.04 (8)} & \paperrev{0.51$\pm$0.05 (10)} & \paperrev{0.50$\pm$0.04 (12)} & \paperrev{0.49$\pm$0.01 (14)} & \paperrev{0.71$\pm$0.06 (5)}\\
ElNino & \paperrev{0.97$\pm$0.01 (6)} & \paperrev{0.50$\pm$0.01 (13)} & \paperrev{0.51$\pm$0.02 (11)} & 0.90$\pm$0.00 (\paperrev{8}) & \paperrev{0.99$\pm$0.00 (4)}\\
Energy & \paperrev{0.90$\pm$0.02 (7)} & \paperrev{0.56$\pm$0.03 (11)} & \paperrev{0.94$\pm$0.00 (5)} & 0.98$\pm$\paperrev{0.01} (\paperrev{4}) & \paperrev{0.79$\pm$0.03 (10)}\\
FishWeight & \paperrev{0.77$\pm$0.06 (6)} & \paperrev{0.49$\pm$0.11 (14)} & \paperrev{0.51$\pm$0.11 (13)} & \paperrev{0.56$\pm$0.10 (8)} & \paperrev{0.67$\pm$0.05 (7)}\\
ForestFires & 1.00$\pm$0.00 (\paperrev{3}) & \paperrev{0.70$\pm$0.03 (13)} & \paperrev{0.74$\pm$0.04 (12)} & 0.96$\pm$0.01 (\paperrev{8}) & \textbf{1.00}$\pm$0.00 (1)\\
GasEmission & \paperrev{0.93$\pm$0.01 (7)} & \paperrev{0.50$\pm$0.02 (12)} & \paperrev{0.51$\pm$0.03 (11)} & \paperrev{0.77}$\pm$0.01 (8) & \paperrev{0.97$\pm$0.00 (5)}\\
HeartFailure & 0.98$\pm$\paperrev{0.00} (\paperrev{4}) & \paperrev{0.56$\pm$0.07 (11)} & \paperrev{0.68$\pm$0.06 (10)} & \paperrev{0.88$\pm$0.01 (7)} & \textbf{1.00}$\pm$0.00 (1)\\
Hepatitis & 1.00$\pm$0.00 (\paperrev{4}) & \paperrev{0.86$\pm$0.02 (12)} & \paperrev{0.85$\pm$0.02 (13)} & 0.96$\pm$0.00 (\paperrev{10}) & \textbf{1.00}$\pm$0.00 (1)\\
IndianLiver & 1.00$\pm$0.00 (\paperrev{3}) & \paperrev{0.70$\pm$0.04 (12)} & \paperrev{0.80$\pm$0.05 (11)} & 0.96$\pm$0.01 (\paperrev{8}) & \textbf{1.00}$\pm$0.00 (1)\\
Maintenance & \paperrev{0.70$\pm$0.02 (7)} & \paperrev{0.50$\pm$0.03 (14)} & \paperrev{0.53$\pm$0.03 (9)} & 0.79$\pm$0.01 (\paperrev{5}) & \paperrev{0.70$\pm$0.05 (6)}\\
Parkinson & \paperrev{0.60$\pm$0.03 (9)} & \paperrev{0.49$\pm$0.03 (13)} & \paperrev{0.50$\pm$0.03 (12)} & \paperrev{0.77$\pm$0.03 (6)} & \paperrev{0.62$\pm$0.03 (8)}\\
PowerPlant & \paperrev{0.80$\pm$0.02 (6)} & 0.50$\pm$0.03 (12) & \paperrev{0.50$\pm$0.02 (11)} & \paperrev{0.63$\pm$0.02 (8)} & 0.71$\pm$0.04 (7)\\
QSRanking & \paperrev{0.73$\pm$0.05 (7)} & \paperrev{0.48$\pm$0.07 (12)} & \paperrev{0.48$\pm$0.04 (13)} & \paperrev{0.52$\pm$0.04 (9)} & \paperrev{0.86$\pm$0.03 (5)}\\
SynMachine & \paperrev{0.85$\pm$0.03 (11)} & 0.92$\pm$0.03 (\paperrev{7}) & 0.99$\pm$\paperrev{0.00} (3) & 0.85$\pm$0.03 (\paperrev{10}) & 0.68$\pm$\paperrev{0.03} (\paperrev{14})\\
Toxicity & \paperrev{0.69$\pm$0.03 (8)} & \paperrev{0.60$\pm$0.05 (11)} & \paperrev{0.59$\pm$0.03 (12)} & \paperrev{0.72$\pm$0.04 (7)} & 0.69$\pm$0.03 (\paperrev{9})\\
Yacht & \paperrev{0.79$\pm$0.04 (6)} & \paperrev{0.71$\pm$0.04 (11)} & \paperrev{0.83$\pm$0.03 (5)} & \paperrev{0.60$\pm$0.08 (14)} & \paperrev{0.74$\pm$0.05 (9)}\\
\midrule
\textbf{Avg Rank} & \paperrev{6.650} & \paperrev{11.100} & \paperrev{10.000} & \paperrev{8.350} & \paperrev{6.250}\\
\midrule
Dataset & LUNAR & DTE-C & \multicolumn{2}{c}{DTE-NP} & TCCM \\
\midrule
Abalone & \paperrev{1.00}$\pm$0.00 (\paperrev{6}) & \paperrev{1.00}$\pm$\paperrev{0.00} (\paperrev{7}) & \multicolumn{2}{c}{\underline{\paperrev{1.00}}$\pm$0.00 (\paperrev{2})} & \paperrev{0.82$\pm$0.03 (13)}\\
Airfoil & \underline{\paperrev{0.83}}$\pm$\paperrev{0.04} (\paperrev{2}) & 0.50$\pm$\paperrev{0.01} (14) & \multicolumn{2}{c}{\paperrev{0.53}$\pm$\paperrev{0.03} (\paperrev{9})} & \paperrev{0.51$\pm$0.05 (12)}\\
BodyFat & \paperrev{0.86$\pm$0.05 (5)} & \paperrev{0.50$\pm$0.09 (13)} & \multicolumn{2}{c}{\paperrev{0.51$\pm$0.06 (12)}} & \paperrev{0.54$\pm$0.07 (9)}\\
Boston & \paperrev{0.69}$\pm$0.03 (5) & \paperrev{0.47$\pm$0.05 (14)} & \multicolumn{2}{c}{\paperrev{0.48$\pm$0.04 (11)}} & \paperrev{0.48$\pm$0.07 (10)}\\
Concrete & \paperrev{0.68$\pm$0.09 (6)} & \paperrev{0.50$\pm$0.00 (13)} & \multicolumn{2}{c}{\paperrev{0.51$\pm$0.03 (11)}} & \paperrev{0.51$\pm$0.05 (9)}\\
ElNino & \paperrev{0.71$\pm$0.02 (9)} & 0.50$\pm$0.00 (\paperrev{14}) & \multicolumn{2}{c}{\paperrev{0.52$\pm$0.02 (10)}} & \paperrev{0.51$\pm$0.02 (12)}\\
Energy & \paperrev{0.85$\pm$0.02 (8)} & 0.50$\pm$0.00 (13) & \multicolumn{2}{c}{\paperrev{0.93}$\pm$\paperrev{0.01} (\paperrev{6})} & \paperrev{0.53$\pm$0.05 (12)}\\
FishWeight & \paperrev{0.84$\pm$0.03 (4)} & \paperrev{0.52$\pm$0.08 (11)} & \multicolumn{2}{c}{\paperrev{0.52$\pm$0.09 (10)}} & \paperrev{0.55$\pm$0.09 (9)}\\
ForestFires & \paperrev{0.94$\pm$0.09 (9)} & \paperrev{0.97}$\pm$\paperrev{0.01} (\paperrev{7}) & \multicolumn{2}{c}{\paperrev{0.78}$\pm$\paperrev{0.02} (\paperrev{11})} & \paperrev{0.99$\pm$0.00 (5)}\\
GasEmission & \paperrev{0.97$\pm$0.05 (6)} & \paperrev{0.50$\pm$0.02 (13)} & \multicolumn{2}{c}{0.52$\pm$\paperrev{0.03} (\paperrev{10})} & \paperrev{0.70$\pm$0.06 (9)}\\
HeartFailure & \paperrev{0.92}$\pm$0.01 (\paperrev{6}) & \paperrev{0.52$\pm$0.05 (12)} & \multicolumn{2}{c}{\paperrev{0.71$\pm$0.06 (9)}} & \paperrev{0.80$\pm$0.06 (8)}\\
Hepatitis & \paperrev{0.98}$\pm$\paperrev{0.01} (9) & \paperrev{0.99$\pm$0.00 (5)} & \multicolumn{2}{c}{\paperrev{0.89}$\pm$\paperrev{0.01} (\paperrev{11})} & \paperrev{0.98$\pm$0.02 (7)}\\
IndianLiver & \paperrev{0.99$\pm$0.01 (5)} & \paperrev{0.58$\pm$0.08 (13)} & \multicolumn{2}{c}{\paperrev{0.87$\pm$0.03 (10)}} & \paperrev{0.96$\pm$0.03 (9)}\\
Maintenance & \textbf{1.00}$\pm$0.00 (\paperrev{1}) & \paperrev{0.51$\pm$0.03 (12)} & \multicolumn{2}{c}{\paperrev{0.57$\pm$0.03 (8)}} & \paperrev{0.53$\pm$0.03 (10)}\\
Parkinson & \paperrev{0.94$\pm$0.01 (4)} & 0.50$\pm$0.00 (11) & \multicolumn{2}{c}{\paperrev{0.53$\pm$0.03 (10)}} & \paperrev{0.71$\pm$0.07 (7)}\\
PowerPlant & \paperrev{0.89$\pm$0.17 (5)} & \paperrev{0.50$\pm$0.01 (13)} & \multicolumn{2}{c}{\paperrev{0.51$\pm$0.02 (10)}} & \paperrev{0.59$\pm$0.13 (9)}\\
QSRanking & \paperrev{0.77$\pm$0.04 (6)} & \paperrev{0.51$\pm$0.05 (10)} & \multicolumn{2}{c}{\paperrev{0.49}$\pm$\paperrev{0.04} (\paperrev{11})} & \paperrev{0.57}$\pm$\paperrev{0.04} (\paperrev{8})\\
SynMachine & \paperrev{0.95}$\pm$\paperrev{0.02} (\paperrev{6}) & \paperrev{0.87}$\pm$0.02 (\paperrev{9}) & \multicolumn{2}{c}{\underline{\paperrev{1.00}}$\pm$0.00 (2)} & \paperrev{0.98}$\pm$\paperrev{0.01} (\paperrev{5})\\
Toxicity & \paperrev{0.76$\pm$0.03 (6)} & \paperrev{0.79}$\pm$\paperrev{0.03} (\paperrev{5}) & \multicolumn{2}{c}{\paperrev{0.63$\pm$0.03 (10)}} & \paperrev{0.51$\pm$0.03 (14)}\\
Yacht & \paperrev{0.76$\pm$0.04 (8)} & \paperrev{0.70$\pm$0.05 (12)} & \multicolumn{2}{c}{0.85$\pm$\paperrev{0.02} (4)} & \paperrev{0.63$\pm$0.02 (13)}\\
\midrule
\textbf{Avg Rank} & \paperrev{5.800} & \paperrev{11.050} & \multicolumn{2}{c}{\paperrev{8.850}} & \paperrev{9.500}\\
\bottomrule

\end{tabular}
}
\end{table}

\begin{table}[tbp]
\centering
\caption{\paperrev{P@$n$ on $20$ datasets. Notation follows Table~\ref{tab:main_results_pr}.}}
\label{tab:main_results_pn}
\scriptsize
\setlength{\tabcolsep}{4pt}
\resizebox{\columnwidth}{!}{
\begin{tabular}{lccccc}
\toprule
Dataset & Flow-CAD (ours) & QCAD & LoPAD & ROCOD & CAD \\
\midrule
Abalone & \paperrev{\textbf{0.97}$\pm$0.02 (1)} & \paperrev{0.91$\pm$0.01 (6)} & \paperrev{0.00}$\pm$0.00 (14) & \paperrev{0.57$\pm$0.01 (10)} & \paperrev{0.94}$\pm$\paperrev{0.01} (\paperrev{3})\\
Airfoil & \paperrev{0.43$\pm$0.07 (3)} & \paperrev{\textbf{0.67}$\pm$0.03 (1)} & \paperrev{0.03$\pm$0.02 (14)} & \paperrev{0.41$\pm$0.04 (4)} & \paperrev{\underline{0.45}$\pm$0.04 (2)}\\
BodyFat & \paperrev{\underline{0.74}$\pm$0.11 (2)} & \paperrev{0.61}$\pm$\paperrev{0.06} (\paperrev{4}) & \paperrev{0.06$\pm$0.06 (14)} & \paperrev{0.70$\pm$0.06 (3)} & \paperrev{\textbf{0.77}$\pm$0.08 (1)}\\
Boston & \paperrev{\underline{0.60}$\pm$0.09 (2)} & \paperrev{\textbf{0.64}$\pm$0.07 (1)} & \paperrev{0.08$\pm$0.01 (8.5)} & \paperrev{0.36$\pm$0.04 (4)} & \paperrev{0.39$\pm$0.08 (3)}\\
Concrete & \paperrev{\textbf{0.61}$\pm$0.11 (1)} & \paperrev{\underline{0.61}$\pm$0.08 (2)} & \paperrev{0.11$\pm$0.04 (7)} & \paperrev{0.33$\pm$0.05 (3)} & \paperrev{0.31$\pm$0.06 (4)}\\
ElNino & \paperrev{0.73$\pm$0.04 (3)} & \textbf{\paperrev{0.94}}$\pm$\paperrev{0.01} (1) & \paperrev{0.12$\pm$0.03 (7)} & \paperrev{0.40$\pm$0.05 (6)} & \paperrev{\underline{0.91}$\pm$0.02 (2)}\\
Energy & \paperrev{\textbf{0.88}$\pm$0.04 (1)} & \paperrev{0.83$\pm$0.03 (3)} & \paperrev{0.00$\pm$0.01 (14)} & \paperrev{0.43$\pm$0.05 (7)} & \paperrev{\underline{0.85}$\pm$0.04 (2)}\\
FishWeight & \paperrev{\textbf{0.83}$\pm$0.08 (1)} & \paperrev{\underline{0.76}$\pm$0.11 (2)} & \paperrev{0.12$\pm$0.10 (9)} & \paperrev{0.41$\pm$0.07 (4)} & \paperrev{0.64$\pm$0.11 (3)}\\
ForestFires & \paperrev{0.91}$\pm$\paperrev{0.05} (4) & \underline{\paperrev{0.95}}$\pm$\paperrev{0.03} (2) & \paperrev{0.10$\pm$0.03 (14)} & \paperrev{0.40$\pm$0.03 (10)} & \paperrev{0.88$\pm$0.02 (5)}\\
GasEmission & \underline{\paperrev{0.92}}$\pm$\paperrev{0.01} (\paperrev{2}) & \paperrev{\textbf{0.94}$\pm$0.01 (1)} & 0.00$\pm$0.00 (14) & \paperrev{0.84$\pm$0.03 (4)} & \paperrev{0.86$\pm$0.02 (3)}\\
HeartFailure & \underline{\paperrev{0.90}}$\pm$0.02 (\paperrev{2}) & \paperrev{0.84$\pm$0.06 (3)} & 0.00$\pm$0.00 (14) & \paperrev{0.15$\pm$0.13 (11)} & \paperrev{0.72$\pm$0.06 (5)}\\
Hepatitis & \textbf{\paperrev{0.97}}$\pm$\paperrev{0.01} (\paperrev{1}) & \paperrev{0.91$\pm$0.03 (3)} & \paperrev{0.05$\pm$0.03 (14)} & \paperrev{0.65$\pm$0.05 (8)} & \paperrev{0.81$\pm$0.04 (5)}\\
IndianLiver & \underline{\paperrev{0.90}}$\pm$\paperrev{0.03} (\paperrev{2}) & \paperrev{0.84$\pm$0.03 (4)} & \paperrev{0.05$\pm$0.03 (14)} & \paperrev{0.51$\pm$0.05 (8)} & \paperrev{0.76$\pm$0.11 (6)}\\
Maintenance & \paperrev{0.70$\pm$0.09 (4)} & \paperrev{\textbf{0.92}$\pm$0.02 (1)} & \paperrev{0.01$\pm$0.01 (8.5)} & \paperrev{0.01$\pm$0.01 (12.5)} & \paperrev{0.85$\pm$0.04 (3)}\\
Parkinson & \textbf{\paperrev{0.94}}$\pm$\paperrev{0.03} (1) & \paperrev{\underline{0.86}$\pm$0.03 (2)} & \paperrev{0.02$\pm$0.02 (7)} & \paperrev{0.55$\pm$0.05 (4)} & \paperrev{0.71$\pm$0.04 (3)}\\
PowerPlant & \paperrev{\textbf{0.71}$\pm$0.04 (1)} & \paperrev{0.49}$\pm$\paperrev{0.04} (4) & 0.01$\pm$0.01 (\paperrev{10}) & \paperrev{\underline{0.69}$\pm$0.04 (2)} & \paperrev{0.53$\pm$0.06 (3)}\\
QSRanking & \paperrev{\underline{0.69}$\pm$0.09 (2.5)} & \textbf{\paperrev{0.71}}$\pm$0.05 (1) & 0.02$\pm$0.02 (14) & \paperrev{0.52$\pm$0.07 (4)} & \paperrev{\underline{0.69}$\pm$0.09 (2.5)}\\
SynMachine & \paperrev{\textbf{1.00}$\pm$0.01 (1)} & \underline{\paperrev{0.96}}$\pm$\paperrev{0.02} (2) & \paperrev{0.41$\pm$0.03 (13)} & \paperrev{0.74$\pm$0.08 (7)} & \paperrev{0.48$\pm$0.11 (11)}\\
Toxicity & \paperrev{\textbf{0.57}$\pm$0.07 (1)} & \paperrev{\underline{0.47}$\pm$0.04 (2)} & \paperrev{0.17$\pm$0.04 (5)} & 0.47$\pm$0.06 (3) & \paperrev{0.46$\pm$0.06 (4)}\\
Yacht & \paperrev{\textbf{0.84}$\pm$0.10 (1)} & \paperrev{\underline{0.81}$\pm$0.05 (2)} & \paperrev{0.56$\pm$0.06 (3)} & \paperrev{0.30$\pm$0.08 (5)} & \paperrev{0.26$\pm$0.06 (9)}\\
\midrule
\textbf{Avg Rank} & \paperrev{\textbf{1.825}} & \paperrev{\underline{2.350}} & \paperrev{10.900} & \paperrev{5.975} & \paperrev{3.975}\\
\midrule
Dataset & IForest & LOF & $k$-NN & SOD & HBOS \\
\midrule
Abalone & \paperrev{0.43$\pm$0.08 (11)} & 0.92$\pm$0.02 (\paperrev{5}) & \paperrev{0.94$\pm$0.01 (4)} & \paperrev{0.74$\pm$0.01 (9)} & \paperrev{0.21$\pm$0.06 (13)}\\
Airfoil & \paperrev{0.12$\pm$0.04 (7.5)} & \paperrev{0.04$\pm$0.03 (12.5)} & \paperrev{0.06$\pm$0.03 (9)} & \paperrev{0.13$\pm$0.03 (6)} & \paperrev{0.12$\pm$0.01 (7.5)}\\
BodyFat & \paperrev{0.19$\pm$0.07 (7)} & \paperrev{0.09$\pm$0.09 (11.5)} & \paperrev{0.11$\pm$0.06 (8)} & \paperrev{0.08$\pm$0.06 (13)} & \paperrev{0.25$\pm$0.06 (6)}\\
Boston & \paperrev{0.11$\pm$0.04 (7)} & \paperrev{0.08$\pm$0.05 (8.5)} & \paperrev{0.06$\pm$0.04 (12)} & \paperrev{0.06$\pm$0.04 (13)} & 0.15$\pm$0.06 (\paperrev{5})\\
Concrete & \paperrev{0.08$\pm$0.03 (8)} & \paperrev{0.07$\pm$0.03 (11)} & \paperrev{0.06$\pm$0.02 (13)} & \paperrev{0.06$\pm$0.02 (12)} & 0.29$\pm$\paperrev{0.06} (5)\\
ElNino & \paperrev{0.53$\pm$0.07 (5)} & \paperrev{0.01$\pm$0.01 (11.5)} & 0.01$\pm$0.01 (\paperrev{11.5}) & \paperrev{0.01$\pm$0.00 (14)} & \paperrev{0.69$\pm$0.03 (4)}\\
Energy & \paperrev{0.49$\pm$0.04 (5)} & \paperrev{0.11$\pm$0.03 (10.5)} & 0.30$\pm$0.03 (\paperrev{9}) & \paperrev{0.80$\pm$0.06 (4)} & \paperrev{0.34$\pm$0.06 (8)}\\
FishWeight & \paperrev{0.23$\pm$0.08 (7)} & \paperrev{0.09$\pm$0.06 (13.5)} & \paperrev{0.10$\pm$0.08 (11)} & 0.09$\pm$0.06 (\paperrev{13.5}) & \paperrev{0.30$\pm$0.15 (6)}\\
ForestFires & 0.94$\pm$\paperrev{0.03} (3) & \paperrev{0.16$\pm$0.05 (12)} & \paperrev{0.13$\pm$0.04 (13)} & 0.60$\pm$0.04 (\paperrev{9}) & \textbf{0.99}$\pm$0.01 (1)\\
GasEmission & \paperrev{0.05$\pm$0.02 (7)} & \paperrev{0.02$\pm$0.01 (10)} & \paperrev{0.01$\pm$0.01 (13)} & \paperrev{0.02$\pm$0.01 (10)} & \paperrev{0.22$\pm$0.05 (6)}\\
HeartFailure & \paperrev{0.80$\pm$0.03 (4)} & \paperrev{0.12$\pm$0.06 (12)} & \paperrev{0.18$\pm$0.06 (10)} & \paperrev{0.32$\pm$0.06 (8)} & \paperrev{\textbf{0.93}$\pm$0.03 (1)}\\
Hepatitis & \paperrev{0.90}$\pm$\paperrev{0.03} (4) & \paperrev{0.07$\pm$0.04 (13)} & \paperrev{0.08$\pm$0.04 (12)} & 0.33$\pm$0.03 (10) & \paperrev{\underline{0.96}$\pm$0.02 (2)}\\
IndianLiver & \paperrev{0.89}$\pm$\paperrev{0.03} (\paperrev{3}) & 0.07$\pm$0.03 (13) & 0.20$\pm$0.10 (\paperrev{11}) & \paperrev{0.41$\pm$0.08 (9)} & \textbf{0.95}$\pm$0.02 (1)\\
Maintenance & \paperrev{0.02}$\pm$\paperrev{0.01} (\paperrev{6}) & 0.00$\pm$0.00 (14) & \paperrev{0.01$\pm$0.00 (12.5)} & \paperrev{0.01$\pm$0.01 (8.5)} & \paperrev{0.36$\pm$0.09 (5)}\\
Parkinson & \paperrev{0.02$\pm$0.01 (8.5)} & \paperrev{0.01$\pm$0.02 (12)} & \paperrev{0.01$\pm$0.01 (14)} & \paperrev{0.02$\pm$0.01 (8.5)} & \paperrev{0.01$\pm$0.01 (11)}\\
PowerPlant & \paperrev{0.08}$\pm$\paperrev{0.03} (8) & \paperrev{0.01$\pm$0.01 (11.5)} & 0.02$\pm$0.02 (\paperrev{9}) & \paperrev{0.01$\pm$0.01 (11.5)} & \paperrev{0.26$\pm$0.04 (6)}\\
QSRanking & \paperrev{0.23$\pm$0.09 (6)} & \paperrev{0.09$\pm$0.05 (8)} & \paperrev{0.08$\pm$0.04 (10)} & \paperrev{0.07$\pm$0.03 (13)} & \paperrev{0.46$\pm$0.06 (5)}\\
SynMachine & \paperrev{0.41$\pm$0.05 (12)} & \paperrev{0.71$\pm$0.05 (8)} & \paperrev{0.79$\pm$0.04 (5)} & \paperrev{0.66$\pm$0.04 (9)} & \paperrev{0.27$\pm$0.04 (14)}\\
Toxicity & \paperrev{0.09$\pm$0.03 (11)} & \paperrev{0.07$\pm$0.02 (12)} & \paperrev{0.05$\pm$0.02 (13)} & \paperrev{0.10$\pm$0.05 (9)} & \paperrev{0.11$\pm$0.03 (8)}\\
Yacht & 0.32$\pm$\paperrev{0.07} (\paperrev{4}) & \paperrev{0.20$\pm$0.05 (13)} & \paperrev{0.23$\pm$0.05 (11.5)} & \paperrev{0.26$\pm$0.06 (8)} & \paperrev{0.29$\pm$0.09 (6)}\\
\midrule
\textbf{Avg Rank} & \paperrev{6.700} & \paperrev{11.125} & \paperrev{10.575} & \paperrev{9.900} & \paperrev{6.025}\\
\midrule
Dataset & LUNAR & DTE-C & \multicolumn{2}{c}{DTE-NP} & TCCM \\
\midrule
Abalone & \paperrev{0.90$\pm$0.02 (7.5)} & \paperrev{0.90$\pm$0.01 (7.5)} & \multicolumn{2}{c}{\paperrev{\underline{0.95}$\pm$0.02 (2)}} & \paperrev{0.32$\pm$0.09 (12)}\\
Airfoil & \paperrev{0.36$\pm$0.03 (5)} & \paperrev{0.04$\pm$0.03 (11)} & \multicolumn{2}{c}{\paperrev{0.06$\pm$0.03 (10)}} & \paperrev{0.04$\pm$0.03 (12.5)}\\
BodyFat & \paperrev{0.44}$\pm$\paperrev{0.09} (5) & \paperrev{0.10$\pm$0.13 (9)} & \multicolumn{2}{c}{\paperrev{0.09$\pm$0.06 (10)}} & \paperrev{0.09$\pm$0.08 (11.5)}\\
Boston & \paperrev{0.14$\pm$0.04 (6)} & \paperrev{0.07$\pm$0.03 (10)} & \multicolumn{2}{c}{\paperrev{0.04$\pm$0.04 (14)}} & \paperrev{0.07$\pm$0.04 (11)}\\
Concrete & \paperrev{0.12$\pm$0.08 (6)} & \paperrev{0.08$\pm$0.02 (9)} & \multicolumn{2}{c}{\paperrev{0.07$\pm$0.02 (10)}} & \paperrev{0.05}$\pm$\paperrev{0.02} (\paperrev{14})\\
ElNino & \paperrev{0.03$\pm$0.01 (8)} & \paperrev{0.01}$\pm$\paperrev{0.01} (\paperrev{10}) & \multicolumn{2}{c}{\paperrev{0.01$\pm$0.00 (9)}} & \paperrev{0.01$\pm$0.01 (13)}\\
Energy & \paperrev{0.46$\pm$0.05 (6)} & \paperrev{0.08$\pm$0.03 (13)} & \multicolumn{2}{c}{\paperrev{0.11$\pm$0.03 (10.5)}} & \paperrev{0.10$\pm$0.04 (12)}\\
FishWeight & \paperrev{0.33$\pm$0.07 (5)} & \paperrev{0.10$\pm$0.09 (11)} & \multicolumn{2}{c}{\paperrev{0.10$\pm$0.05 (11)}} & \paperrev{0.13$\pm$0.08 (8)}\\
ForestFires & \paperrev{0.73$\pm$0.13 (8)} & \paperrev{0.73$\pm$0.07 (7)} & \multicolumn{2}{c}{\paperrev{0.21$\pm$0.03 (11)}} & \paperrev{0.86$\pm$0.05 (6)}\\
GasEmission & \paperrev{0.39$\pm$0.16 (5)} & \paperrev{0.02$\pm$0.01 (12)} & \multicolumn{2}{c}{\paperrev{0.02$\pm$0.01 (10)}} & \paperrev{0.02$\pm$0.01 (8)}\\
HeartFailure & \paperrev{0.46$\pm$0.06 (6)} & \paperrev{0.10$\pm$0.04 (13)} & \multicolumn{2}{c}{\paperrev{0.22$\pm$0.06 (9)}} & \paperrev{0.41$\pm$0.12 (7)}\\
Hepatitis & \paperrev{0.57$\pm$0.11 (9)} & \paperrev{0.78$\pm$0.04 (6)} & \multicolumn{2}{c}{\paperrev{0.08$\pm$0.05 (11)}} & \paperrev{0.70$\pm$0.27 (7)}\\
IndianLiver & \paperrev{0.82$\pm$0.04 (5)} & \paperrev{0.11$\pm$0.06 (12)} & \multicolumn{2}{c}{\paperrev{0.31$\pm$0.05 (10)}} & \paperrev{0.55$\pm$0.11 (7)}\\
Maintenance & \paperrev{\underline{0.87}$\pm$0.09 (2)} & \paperrev{0.01$\pm$0.01 (10.5)} & \multicolumn{2}{c}{\paperrev{0.01$\pm$0.01 (7)}} & \paperrev{0.01$\pm$0.01 (10.5)}\\
Parkinson & \paperrev{0.11$\pm$0.03 (6)} & \paperrev{0.02$\pm$0.01 (10)} & \multicolumn{2}{c}{\paperrev{0.01$\pm$0.01 (13)}} & \paperrev{0.23$\pm$0.14 (5)}\\
PowerPlant & \paperrev{0.47$\pm$0.15 (5)} & \paperrev{0.01}$\pm$0.01 (\paperrev{13.5}) & \multicolumn{2}{c}{\paperrev{0.01$\pm$0.01 (13.5)}} & \paperrev{0.26$\pm$0.13 (7)}\\
QSRanking & \paperrev{0.11}$\pm$\paperrev{0.06} (\paperrev{7}) & \paperrev{0.09$\pm$0.06 (9)} & \multicolumn{2}{c}{\paperrev{0.08$\pm$0.06 (11.5)}} & \paperrev{0.08$\pm$0.05 (11.5)}\\
SynMachine & \paperrev{0.77$\pm$0.05 (6)} & \paperrev{0.55$\pm$0.05 (10)} & \multicolumn{2}{c}{\paperrev{0.93$\pm$0.02 (3)}} & \paperrev{0.82}$\pm$\paperrev{0.04} (\paperrev{4})\\
Toxicity & \paperrev{0.17$\pm$0.04 (6)} & \paperrev{0.13$\pm$0.03 (7)} & \multicolumn{2}{c}{\paperrev{0.05}$\pm$\paperrev{0.01} (\paperrev{14})} & \paperrev{0.09$\pm$0.09 (10)}\\
Yacht & \paperrev{0.23$\pm$0.10 (11.5)} & \paperrev{0.24$\pm$0.08 (10)} & \multicolumn{2}{c}{\paperrev{0.28}$\pm$\paperrev{0.06} (\paperrev{7})} & \paperrev{0.19$\pm$0.05 (14)}\\
\midrule
\textbf{Avg Rank} & \paperrev{6.250} & \paperrev{10.025} & \multicolumn{2}{c}{\paperrev{9.825}} & \paperrev{9.550}\\
\bottomrule

\end{tabular}
}
\end{table}

\subsection{\texorpdfstring{\paperrev{Comparison with Existing Detectors} (RQ2)}{Comparison with Existing Detectors (RQ2)}}

\paperrev{We compare Flow-CAD with contextual and point anomaly detectors, first within each group and then against the strongest competitor for each dataset and metric.}

\paperrev{\textbf{1) Contextual anomaly detectors.} Among QCAD, CAD, ROCOD, LoPAD, and Flow-CAD, Flow-CAD has the best average rank for all three metrics. Its average ranks are $1.900$, $2.050$, and $1.825$, compared with $2.300$, $2.100$, and $2.350$ for QCAD. Flow-CAD has higher PR AUC and P@$n$ than QCAD on $13$ of the $20$ datasets, while the two methods each lead on ten datasets for ROC AUC. Flow-CAD also leads CAD on at least $15$ datasets for every metric, and leads ROCOD and LoPAD on all $20$.}

\paperrev{\textbf{2) Point anomaly detectors.} The best average ranks among the nine point detectors are $6.350$ for PR AUC (HBOS), $5.800$ for ROC AUC (LUNAR), and $6.025$ for P@$n$ (HBOS). Flow-CAD's corresponding ranks are $1.900$, $2.050$, and $1.825$. Across datasets, it leads every point detector in $14$ PR AUC cases, $14$ ROC AUC cases, and $16$ P@$n$ cases.}

\paperrev{\textbf{3) Comparison with the strongest baseline on each table.} The leading baseline changes with the dataset and metric. We therefore compare Flow-CAD with the best member of each group in every case. Across the $60$ dataset--metric combinations, Flow-CAD exceeds the strongest contextual baseline in $34$ cases and the strongest point detector in $44$. Its average-rank advantage is thus supported by results across different datasets and metrics rather than by one favorable subset.}

\subsection{\texorpdfstring{Instance-Level Interpretability via\protect\\Residual Decomposition (RQ3)}{Instance-Level Interpretability via Residual Decomposition (RQ3)}}
\label{sub:exp:XAD}
\paperrev{Flow-CAD assigns an anomaly score to each row and records how much each behavioral column contributes to that score. Figure~\ref{fig:feature_contrib} shows both quantities for one anomalous row from the Energy Efficiency dataset and a nearby normal row observed under a similar context. For this illustration, we train Flow-CAD on the QCAD benchmark table using the original Energy Efficiency configuration.}

\Needspace{5\baselineskip}
\paperrev{\textbf{1) Why the anomalous row receives a high score.} Figure~\ref{fig:feature_contrib}(a) shows sampled points along the paths to the two observed rows. At each point, the arrow is the difference between the velocity predicted by Flow-CAD and the target velocity for that path. The anomalous row repeatedly produces longer and more dispersed arrows than the nearby normal row. Its high score therefore reflects large residuals across many Monte Carlo draws, rather than one unusual sample.}

{\setlength{\intextsep}{4pt plus 1pt minus 1pt}
\begin{figure}[t]
    \centering
    \includegraphics[width=0.98\linewidth]{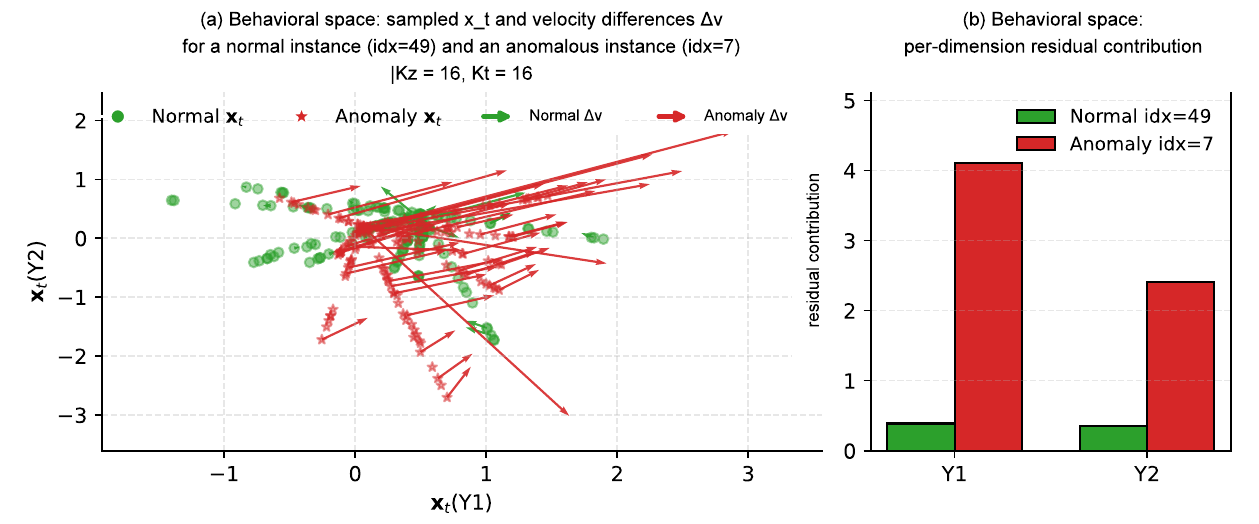}
    \caption{Illustrative raw-score decomposition for two Energy Efficiency instances with behavioral features $(Y_1,Y_2)$. (a) Sampled intermediate states and \paperrev{signed velocity differences $\Delta\boldsymbol v$; the score uses their coordinate-wise magnitudes $\boldsymbol r=|\Delta\boldsymbol v|$}. (b) Feature contributions estimated with \paperrev{a separate set of Monte Carlo draws}.}
    \label{fig:feature_contrib}
\end{figure}}

\paperrev{\textbf{2) Which columns drive the score.} Figure~\ref{fig:feature_contrib}(b) reports the residual contribution of each behavioral column after MAD normalization and truncation. For the anomalous row, $Y_1$ and $Y_2$ contribute about $4.1$ and $2.4$, respectively, whereas both contributions are approximately $0.4$ or less for the normal row. Because the contributions $s_q$ sum to the raw score in~\eqref{eq:flowcad-raw-score}, their relative sizes indicate which column should be inspected first. In this example, the larger contribution of $Y_1$ directs attention to that column when the row is compared with observations from similar contexts.}

\subsection{Ablation and Sensitivity Analysis (RQ4)}
\label{sub:exp:abaltion}

\begin{figure}[tbp]
    \centering
    \includegraphics[width=1\linewidth]{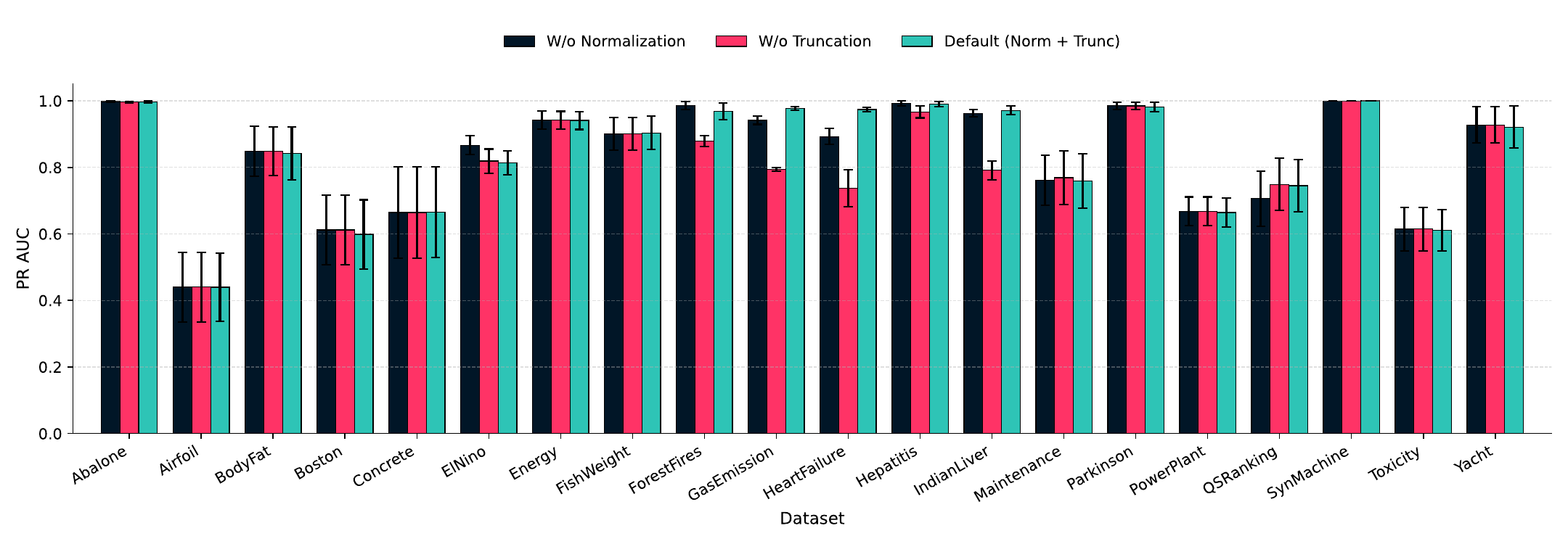}\\[0.8em]
    \includegraphics[width=1\linewidth]{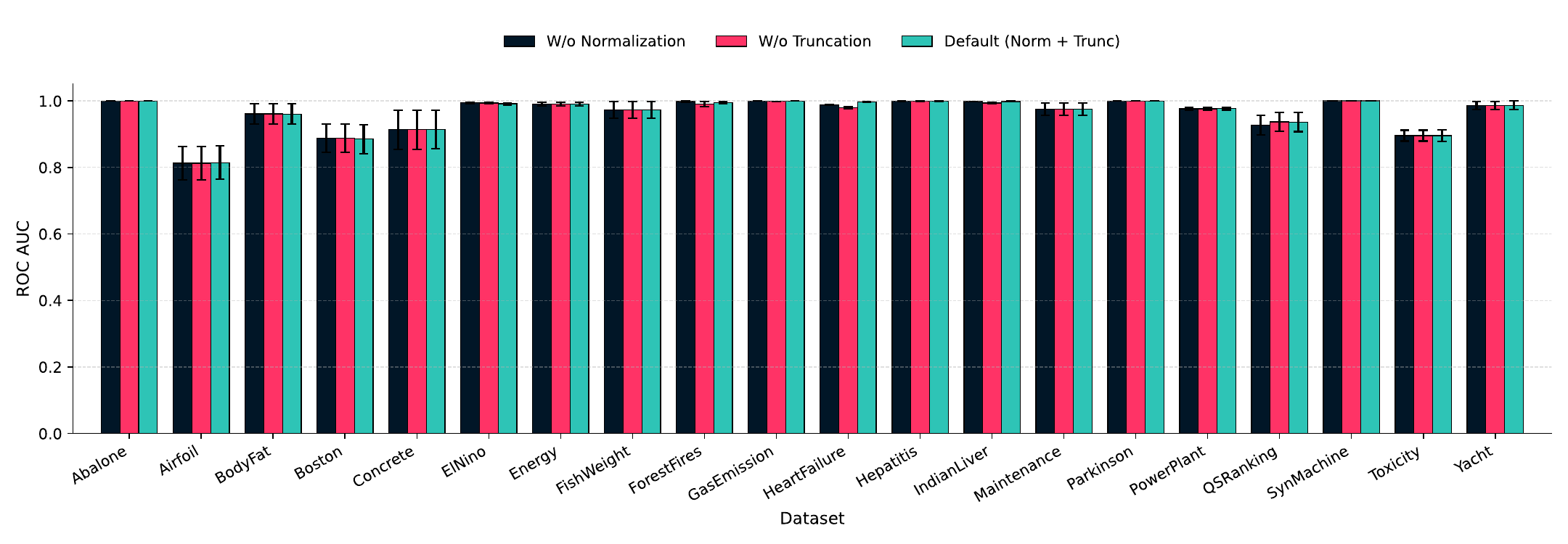}\\[0.8em]
    \includegraphics[width=1\linewidth]{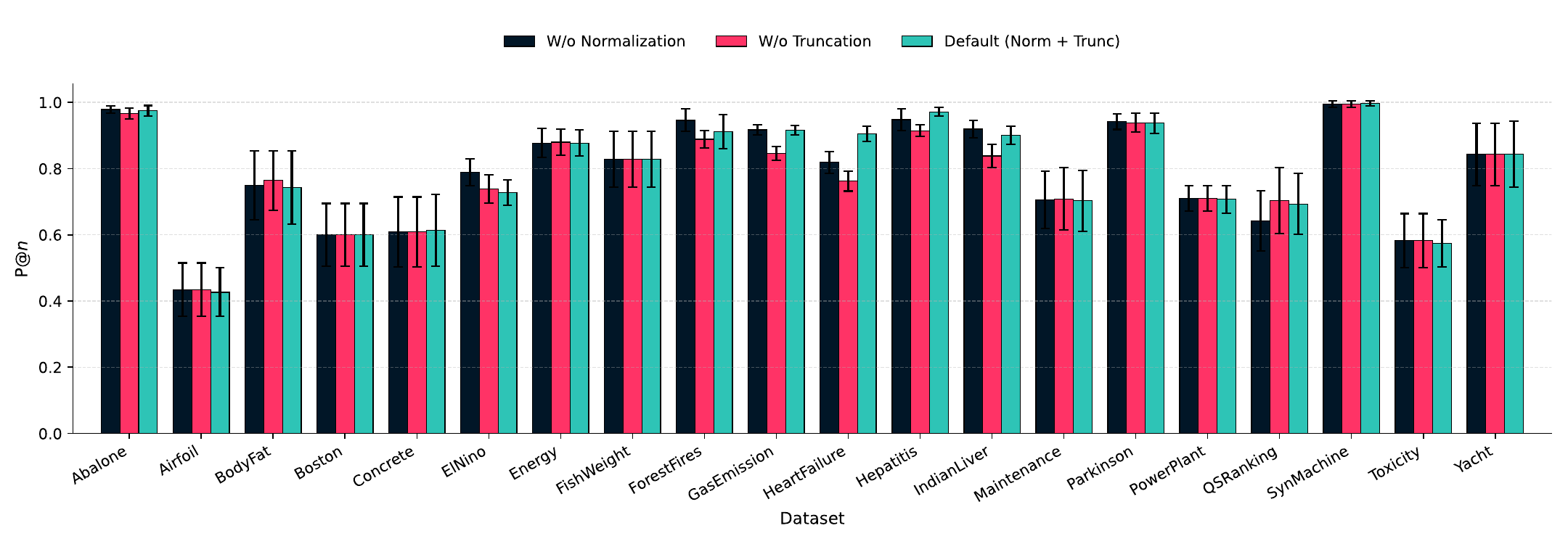}
    \caption{Ablation on residual normalization and truncation across all datasets. Panels show PR AUC, ROC AUC, and P@$n$ (top to bottom). We compare the default configuration (MAD normalization with truncation) against variants without normalization and without truncation.}
    \label{fig:ablation_all}
\end{figure}

We conduct a series of ablation studies to assess the sensitivity of Flow-CAD to key design choices, including residual normalization, truncation, aggregation norms, and Monte Carlo sampling. \paperrev{Figures~\ref{fig:ablation_all}--\ref{fig:sensitivity_kz} vary one scoring choice at a time on the same tables and $1{,}000$-epoch fits as the main benchmark. They report means and standard deviations across the seven benchmark runs, using MAD normalization, $\tau=10$, $L_1$ aggregation, and $K_t=K_z=16$ as the defaults. We also test whether repeated scoring flags the same rows and how the encoding of categorical context affects detection.}

\subsubsection{Ablation studies on residual normalization and truncation}
\label{sec:full-normalization-ablation}
We compare the default configuration (MAD normalization with truncation) against variants that remove either component. \paperrev{Normalization puts residuals from different behavioral columns on a comparable scale, while truncation limits the influence of any single residual. Across the $20$ datasets, the complete score has the highest mean on all three metrics. Truncation provides the larger average gain: removing it lowers mean PR AUC by $3.3$ percentage points and mean P@$n$ by $1.5$ points. MAD normalization gives smaller gains on average, but is important when residual scales differ substantially across columns. On HeartFailure and GasEmission, the complete score gives PR AUCs of $0.974$ and $0.976$, versus $0.893$ and $0.941$ without normalization and $0.738$ and $0.794$ without truncation.}

\subsubsection{Sensitivity to truncation threshold \texorpdfstring{$\tau$}{tau}}
\label{sec:full-truncation-sensitivity}
\begin{figure}[tbp]
    \centering
    \includegraphics[width=1\linewidth]{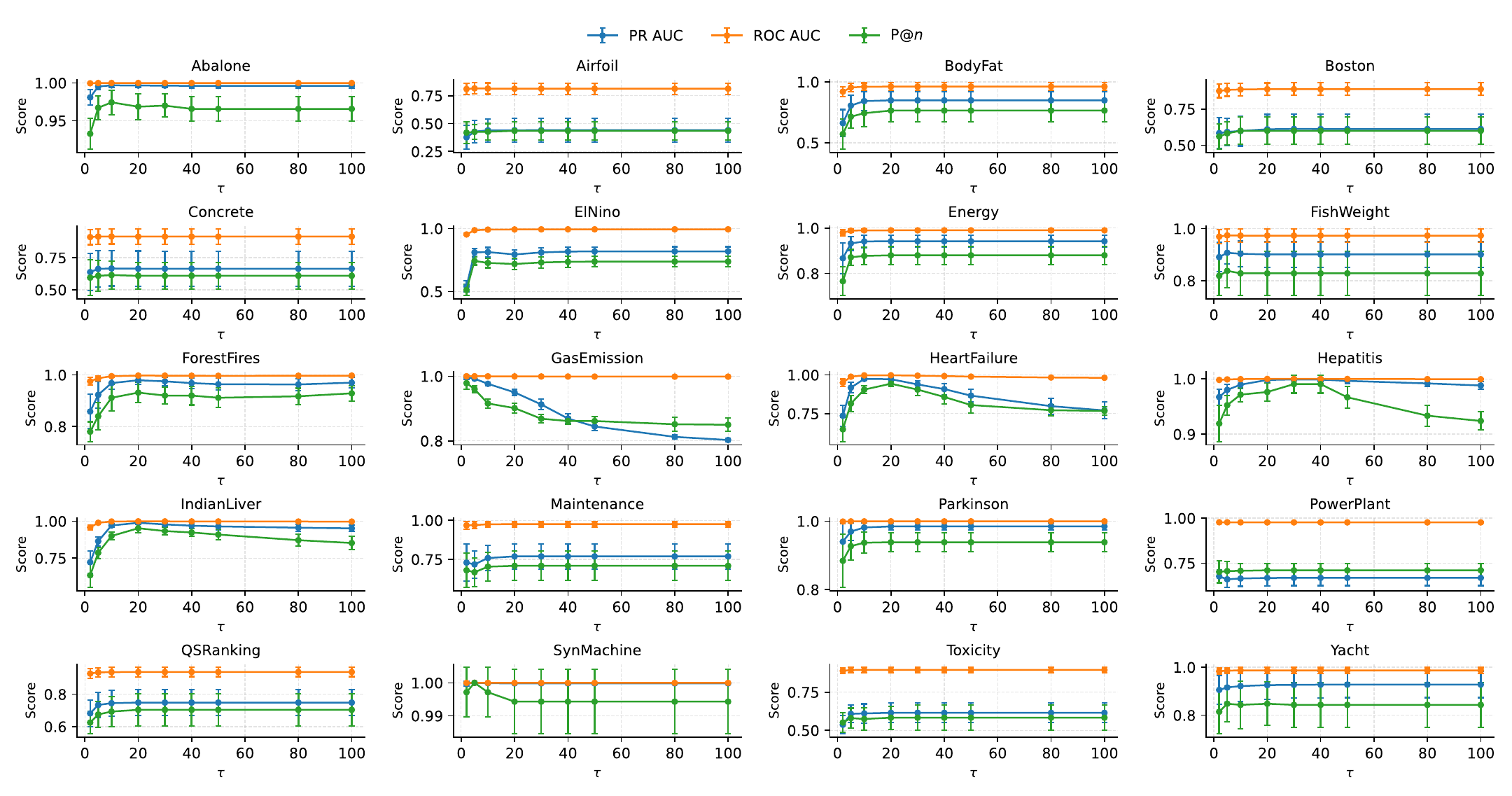}
    \caption{Sensitivity to truncation with MAD normalization. \paperrev{The threshold $\tau$ varies from~$2$~to~$100$.}}
    \label{fig:sensitivity_tau}
\end{figure}

Figure~\ref{fig:sensitivity_tau} varies the \paperrev{positive} truncation threshold $\tau$ \paperrev{from $2$ to $100$} while keeping MAD normalization enabled. \paperrev{Across $\tau\in\{5,10,20\}$, the median per-dataset PR AUC range is only $0.015$. At the default $\tau=10$, PR AUC is within one percentage point of the best of these three settings on $15$ datasets and within two points on all $20$. Thus, $\tau=10$ lies in a broad stable region rather than depending on a narrow optimum. The no-truncation case is reported separately in Figure~\ref{fig:ablation_all}.}

\subsubsection{Sensitivity to residual aggregation norm}
\label{sec:full-aggregation-ablation}
\begin{figure}[tbp]
    \centering
    \includegraphics[width=1\linewidth]{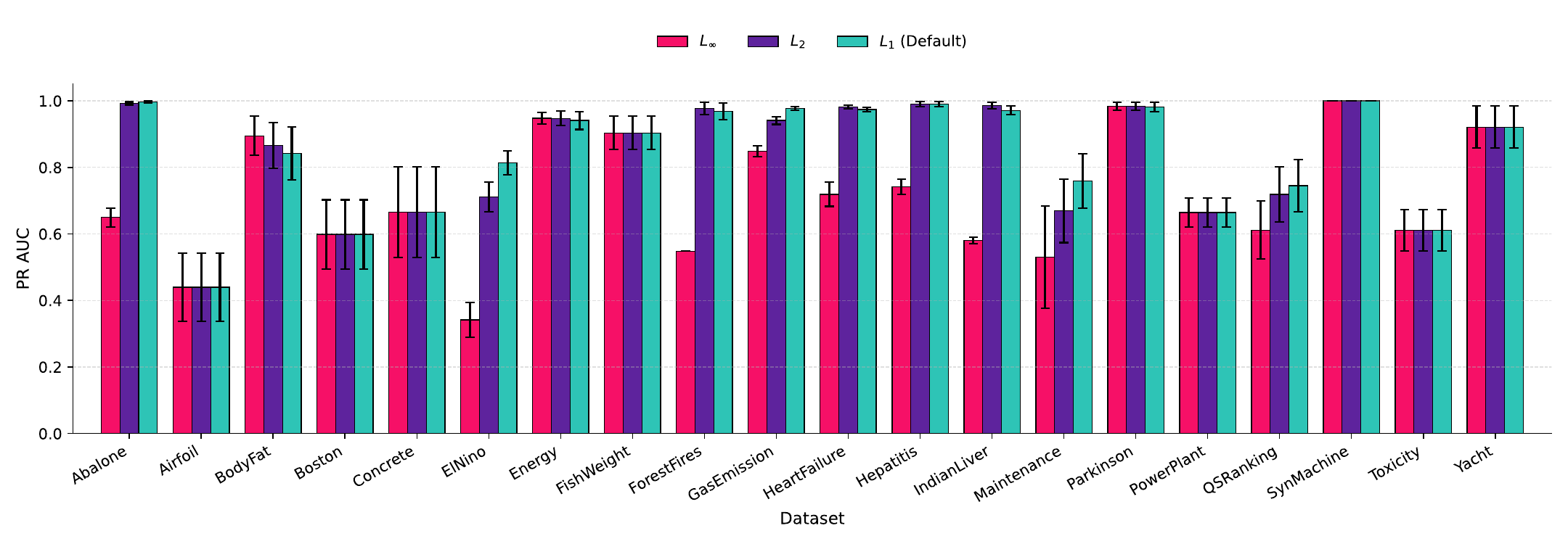}\\[0.8em]
    \includegraphics[width=1\linewidth]{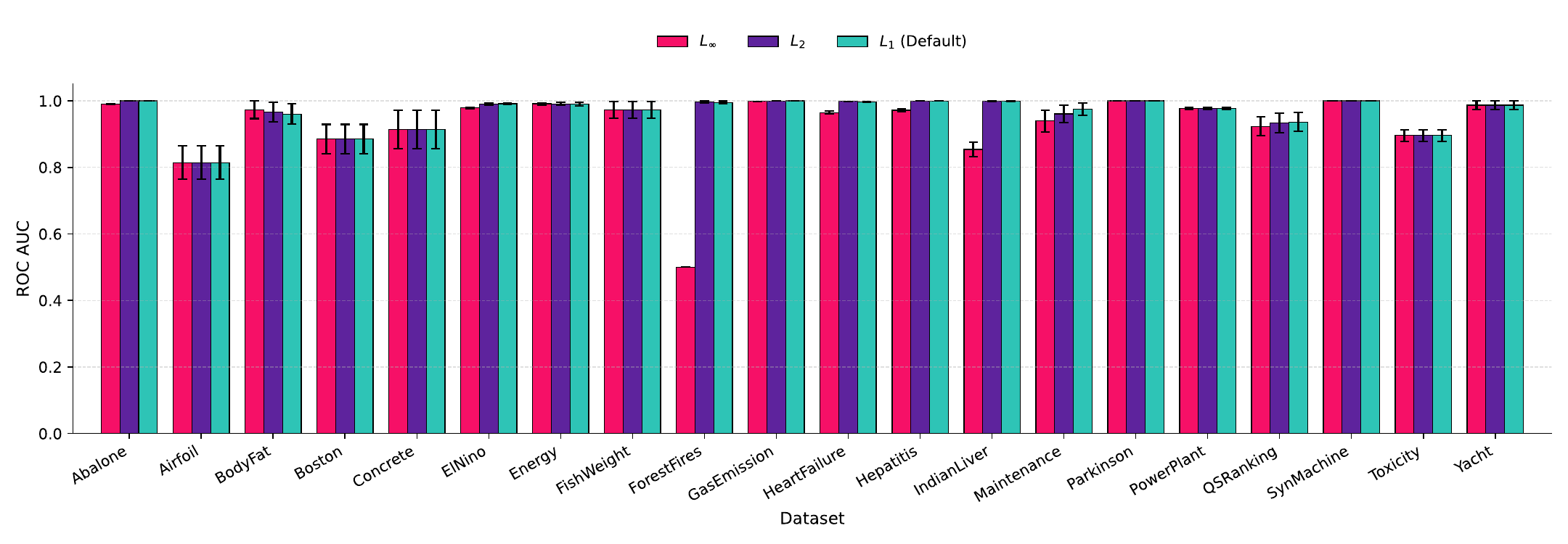}\\[0.8em]
    \includegraphics[width=1\linewidth]{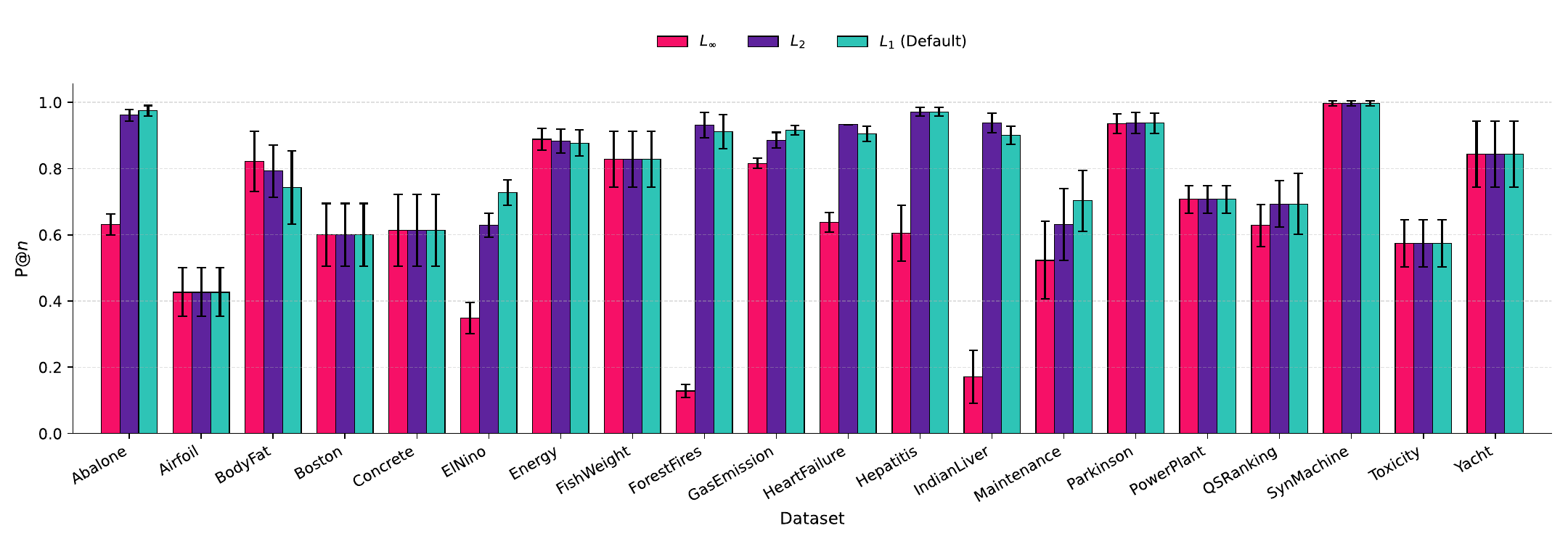}
    \caption{Residual aggregation with $L_1$, $L_2$, and \paperrev{$L_\infty$} norms. Panels show PR AUC, ROC AUC, and P@$n$ (top to bottom).}
    \label{fig:ablation_norm_all}
\end{figure}

Figure~\ref{fig:ablation_norm_all} compares the default $L_1$ norm with the $L_2$ and \paperrev{$L_\infty$} norms. \paperrev{Across the $20$ datasets, $L_1$ has the highest mean for all three metrics: $0.838$ for PR AUC, $0.964$ for ROC AUC, and $0.792$ for P@$n$. The corresponding means are $0.829$, $0.964$, and $0.789$ for $L_2$, and $0.710$, $0.927$, and $0.636$ for $L_\infty$. Moreover, $L_1$ is best or tied on $13$, $13$, and $15$ datasets, respectively, and is within one percentage point of the best norm in $52$ of the $60$ dataset--metric comparisons. We therefore retain $L_1$ as the most reliable overall choice; it also preserves the additive column-wise decomposition used in Section~\ref{sec:scoring}.}

\Needspace{4\baselineskip}
\subsubsection{Sensitivity to Monte Carlo sampling budgets}
\label{sec:full-mc-sensitivity}
\begin{figure}[tbp]
    \centering
    \includegraphics[width=1\linewidth]{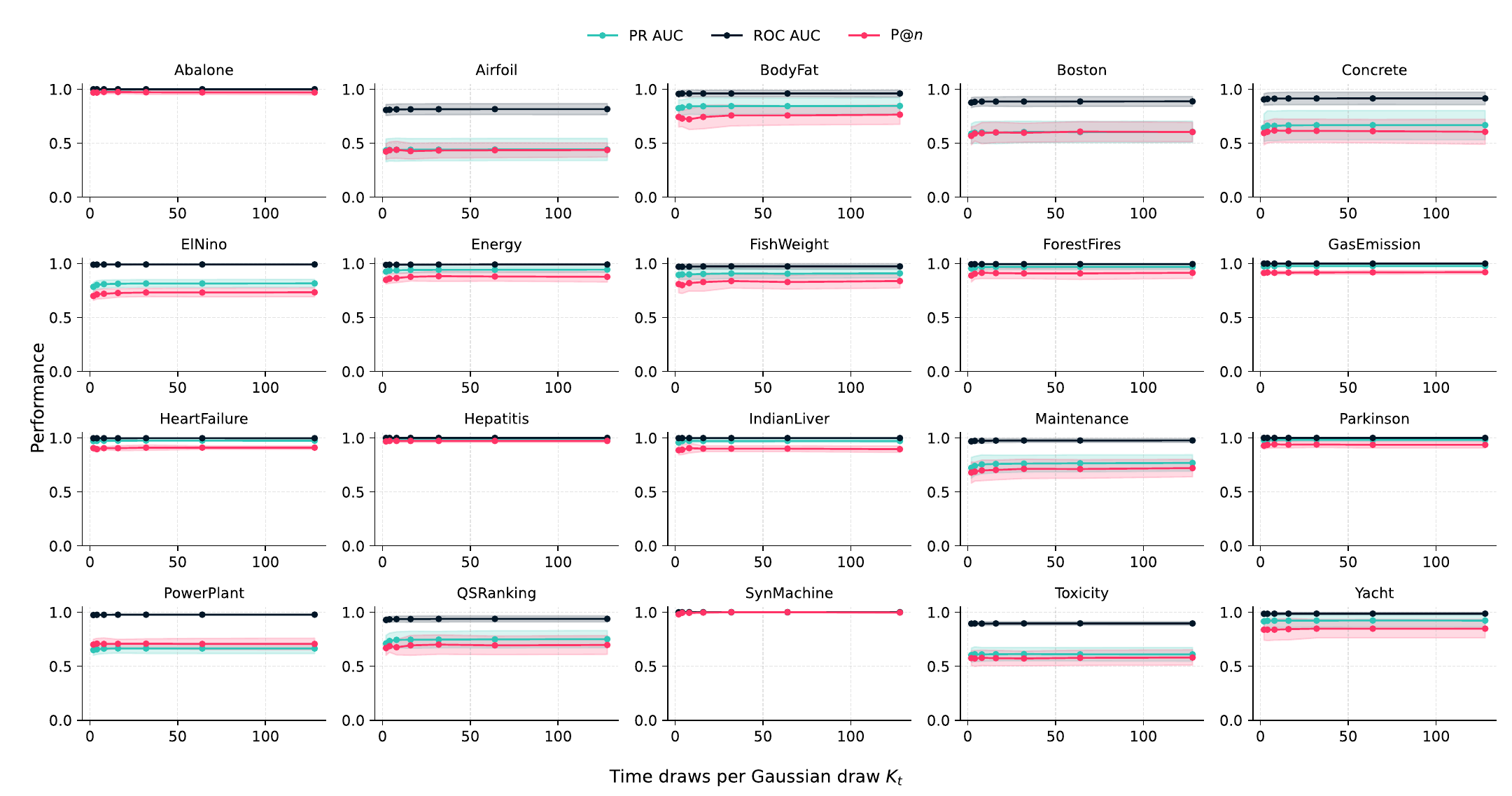}
    \caption{Sensitivity to \paperrev{the number of time draws $K_t$ per Gaussian draw} with MAD~normalization~and~truncation.}
    \label{fig:sensitivity_kt}
\end{figure}

\begin{figure}[tbp]
    \centering
    \includegraphics[width=1\linewidth]{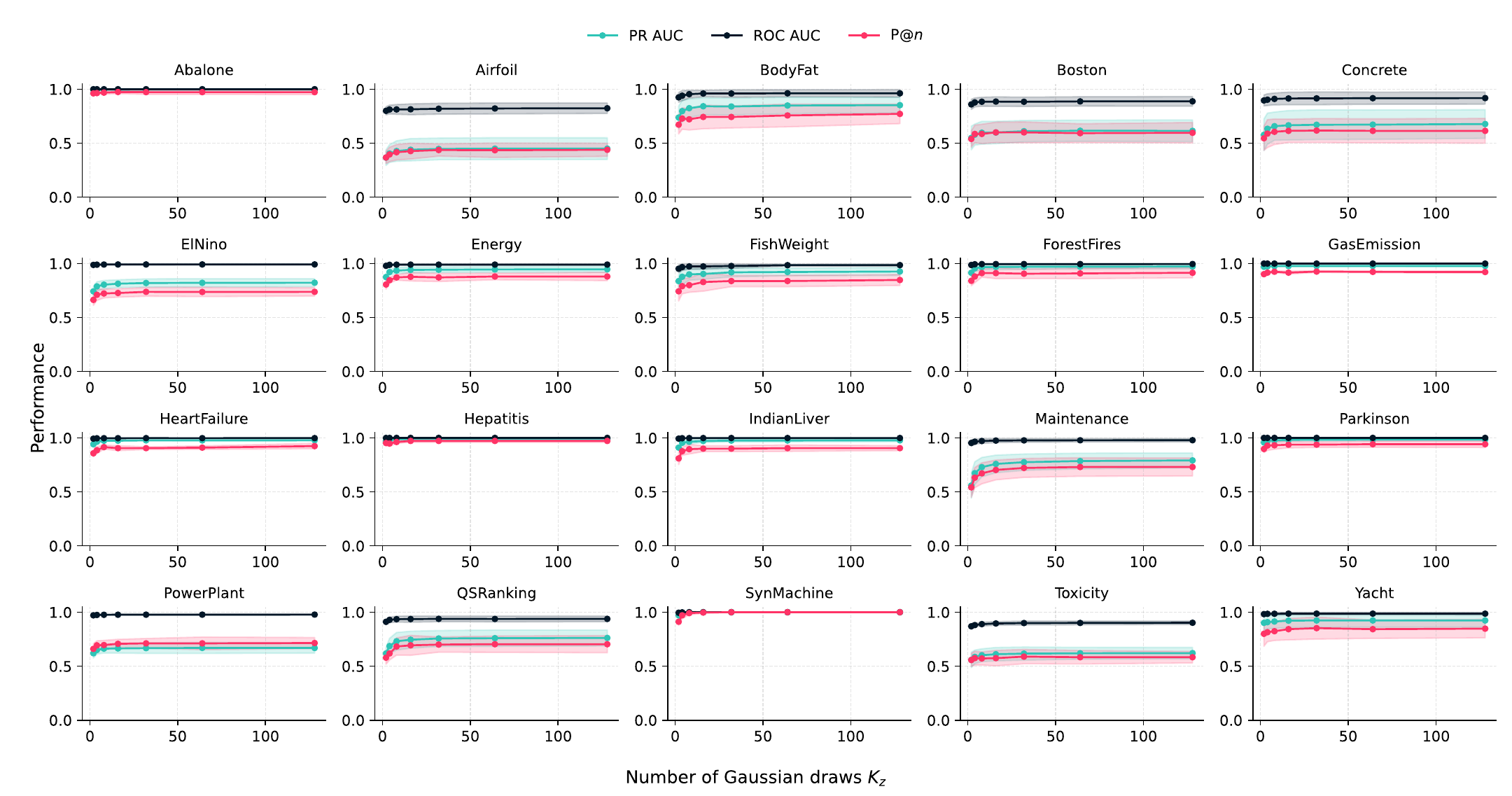}
    \caption{Sensitivity to the number of Gaussian draws $K_z$ with MAD~normalization~and~truncation.}
    \label{fig:sensitivity_kz}
\end{figure}

\paperrev{Figures~\ref{fig:sensitivity_kt} and~\ref{fig:sensitivity_kz} show the effect of varying $K_t$ (time draws per Gaussian draw) and $K_z$ (Gaussian draws), respectively. We hold one budget at $16$ while varying the other and reuse draws across settings. Residual normalization and truncation remain enabled.}

\paperrev{Both budgets matter most when they are small. Increasing $K_z$ from $2$ to $16$ improves PR AUC on all $20$ datasets, with a median gain of $5.3$ percentage points. Increasing $K_t$ over the same range also improves all $20$, with a median gain of $1.0$ point. The gains then taper: moving from $16$ to $128$ adds median gains of $0.5$ points for $K_z$ and $0.1$ points for $K_t$. The larger effect of $K_z$ agrees with Proposition~\ref{prop:mc-concentration}: additional Gaussian draws reduce both variance components, whereas additional time draws reduce only the inner component.}

\subsubsection{\texorpdfstring{\paperrev{Stability of rows selected for review}}{Stability of rows selected for review}}
\label{sec:review-list-stability}
\paperrev{In practice, repeated runs should flag largely the same rows for inspection. We test this by scoring each table twice with independent Monte Carlo draws and comparing the two top-$5\%$ lists. The experiment covers six datasets, each with three single-column variants, for a total of $18$ fitted models. We fix $K_t=16$ and increase $K_z$ from $16$ to $64$. The median overlap rises from $80.4\%$ to $91.2\%$, and $17$ of the $18$ models show higher overlap. More draws therefore make row selection more stable.}

\Needspace{6\baselineskip}
\subsubsection{\texorpdfstring{\paperrev{Representing categorical context}}{Representing categorical context}}
\paperrev{Categorical values such as countries have no natural numerical order, yet scalar codes in $[0,1]$ impose one. We compare scalar and one-hot encodings on FishWeight and QSRanking using the same seven benchmark runs (data seeds $43$--$49$) and $1{,}000$ training epochs. One-hot encoding performs better on all three metrics in both datasets. On FishWeight, PR AUC, ROC AUC, and P@$n$ rise from $0.859$, $0.955$, and $0.781$ to $0.904$, $0.973$, and $0.829$. On QSRanking, they rise from $0.693$, $0.911$, and $0.646$ to $0.745$, $0.936$, and $0.693$. QSRanking has $57$ country categories, so one-hot encoding is the more natural choice.}

\Needspace{6\baselineskip}
\subsection{\texorpdfstring{\paperrev{Controlled Sign-Parity Experiment}}{Controlled Sign-Parity Experiment}}
\label{sec:conditional-diagnostic}

\paperrev{Results on real datasets do not show whether a detector has learned the joint relationship among behavioral columns or is relying on simpler one- or two-column patterns. We therefore construct a sign-parity task in which the context mismatch appears only in the joint pattern of three columns. Any one or two columns have the same distribution in both contexts, so success requires modeling all three together.}

\paperrev{\textbf{1) Test construction and evaluation.} Each block contains all eight sign patterns for a fixed set of absolute values, and the parity of the negative signs determines the correct context. We train for $1{,}000$ epochs on $800$ rows, assigning the wrong context to $5\%$ of them. The seven runs pair data seeds $43$--$49$ with training seeds $202$--$208$. We then score $800$ new vectors under both the correct and wrong contexts. A pair earns one point when the wrong context receives a higher anomaly score and half a point for a tie. For each fitted flow, both contexts and all eight sign patterns in a test block share the same draws.}

\paperrev{\textbf{2) Comparison methods.} The conditional-mean MLP tests whether mean prediction is sufficient; it uses the same encoder, hidden widths, initialization seed, and training MAD scales as Flow-CAD. Coordinatewise Flow fits one flow per column and adds their scores. Zero-context Flow keeps the joint model but removes context by setting its context inputs to zero.}

\paperrev{\textbf{3) Why separate column scores are insufficient.} Fix one sign within a block. Across the four remaining patterns, each column takes the same values in both contexts. For any method that adds separate column scores, the four score gaps (wrong context minus correct context) must therefore sum to zero. If three gaps exceed the tie tolerance, the fourth must favor the correct context and count as a loss. Even with half credit for ties, paired accuracy cannot exceed $75\%$.}

\paperrev{\textbf{4) Results.} Flow-CAD gives the wrong context a higher anomaly score for every test pair in all seven runs, achieving $100\%$ paired accuracy (Table~\ref{tab:conditional-diagnostic}). The three alternatives remain near $50\%$. Flow-CAD gains $53.6$ percentage points over the conditional-mean MLP (95\% paired bootstrap interval across seeds: $[50.0,60.7]$). The joint flow therefore detects a three-column mismatch that is not captured by mean prediction, separate column scores, or a joint model without context.}

\begin{table}[!ht]
    \centering
    \caption{\paperrev{Mean paired accuracy (\%) over seven sign-parity runs.}}
    \label{tab:conditional-diagnostic}
    \scriptsize
    \begingroup\revisioncolor
    \begin{tabular}{lr}
        \toprule
        Method & Paired accuracy \\
        \midrule
        Flow-CAD & 100.00 \\
        Conditional-mean MLP & 46.43 \\
        Coordinatewise Flow & 49.98 \\
        Zero-context Flow & 50.00 \\
        \bottomrule
    \end{tabular}
    \endgroup
\end{table}

\Needspace{5\baselineskip}
\subsection{Runtime Comparison}
\label{sec:runtime-comparison}

\paperrev{We compare Flow-CAD and QCAD on tables of different sizes, using one Apple M4 CPU thread per run. Scoring time covers row scores and column contributions. Total time also covers data loading, fitting or setup, and metric computation.}

\paperrev{\textbf{Effect of table size.} To isolate the effect of row count on runtime, we test four nested PowerPlant subsets from $1{,}000$ to $8{,}000$ rows at three seeds ($12$ paired runs), keeping the columns fixed. QCAD uses $500$ neighbors; Flow-CAD trains for $1{,}000$ epochs. Flow-CAD is faster in every run. At $8{,}000$ rows, its median scoring and total times are $2.57$ and $38.33$ CPU seconds; QCAD takes $54.88$ and $55.99$ seconds, respectively.}

\paperrev{\textbf{Timing on the full benchmark.} To see if the PowerPlant result holds elsewhere, we compare all $20$ benchmarks ($159$ to $20{,}000$ rows). Flow-CAD scores each table $4.0\times$ to $164.1\times$ faster and has lower total time on $16$ of them. On ElNino, total time falls from $595$ CPU seconds with QCAD to $69$ with Flow-CAD. Flow-CAD reuses one fitted model across rows; QCAD builds a local neighborhood and estimates conditional quantiles for each~row.}

\paperrev{For each row, Flow-CAD evaluates $K_z$ Gaussian draws at $K_t$ time points. With fixed network widths, scoring $N$ rows with $P$ encoded context columns and $Q$ behavioral columns requires $\mathcal O(NK_zK_t(P+Q))$ operations, while storing the draws requires $\mathcal O(NK_z(Q+K_t))$~memory.}

\paperrev{\textbf{Scaling to large tables.} Since the benchmarks stop at $20{,}000$ rows, we also test larger inputs by fitting Flow-CAD on $8{,}000$ synthetic rows and scoring in $1{,}000$-row batches with the training MAD scales. It scores one million rows with $64$ encoded context and $10$ behavioral columns in $564$ CPU seconds. With $256$ context and $128$ behavioral columns, it scores $100{,}000$ rows in $228$ seconds. Peak memory is $322$ and $364$ MiB, respectively; batch scoring does not raise the peak reached during fitting.}

\Needspace{5\baselineskip}
\section{Conclusion}

We proposed \textbf{Flow-CAD}, a contextual anomaly detection framework for tabular data based on Flow Matching. Flow-CAD learns a single global and amortized model and employs a residual-based scoring mechanism that avoids explicit likelihood evaluation while providing native per-feature explanations. Experiments on $20$ real-world datasets show that Flow-CAD \paperrev{has the best average rank for PR AUC, ROC AUC, and P@$n$}. \paperrev{The ablations show stable performance around the default truncation and sampling settings, while the timing study shows faster scoring than QCAD on every benchmark table.} These results demonstrate the effectiveness and practicality of Flow Matching for contextual anomaly detection. Future work will explore extensions to structured~and~temporal~contexts.

\section*{Acknowledgment}

\paperrev{This publication is part of the project Digital Twin with project number P18-03 of the research programme TTW Perspective, which is (partly) financed by the Dutch~Research~Council~(NWO).}

\paperrev{We used GPT for language editing and for programming assistance with running experiments.}

\bibliographystyle{IEEEtranN}
\begingroup
\expandafter\def\csname revisedref:ang2026rfod\endcsname{}
\expandafter\def\csname revisedref:nalisnick2019deep\endcsname{}
\expandafter\def\csname revisedref:han2022adbench\endcsname{}
\expandafter\def\csname revisedref:tong2024minibatch\endcsname{}
\expandafter\def\csname revisedref:zhao2019pyod\endcsname{}
\expandafter\def\csname revisedref:li2026diffusion\endcsname{}
\expandafter\def\csname revisedref:bindini2026dealing\endcsname{}
\expandafter\def\csname revisedref:chen2026flowmismatching\endcsname{}
\expandafter\def\csname revisedref:konz2026unsupervised\endcsname{}
\expandafter\def\csname revisedref:ding2025cbaclean\endcsname{}
\expandafter\def\csname revisedref:siddiqi2023saga\endcsname{}
\expandafter\def\csname revisedref:schelter2018quality\endcsname{}
\expandafter\def\csname revisedref:kingma2017adammethodstochasticoptimization\endcsname{}
\let\unmarkedbibitem\bibitem
\RenewDocumentCommand{\bibitem}{o m}{%
  \par
  \normalcolor
  \ifcsname revisedref:#2\endcsname\revisioncolor\fi
  \IfNoValueTF{#1}{\unmarkedbibitem{#2}}{\unmarkedbibitem[#1]{#2}}%
}
\bibliography{references}
\endgroup

\end{document}